\documentclass{article}
\PassOptionsToPackage{numbers}{natbib}
\usepackage[main,final]{neurips_2026}

\usepackage[utf8]{inputenc}
\usepackage[T1]{fontenc}
\usepackage{hyperref}
\usepackage{url}
\usepackage{booktabs}
\usepackage{amsmath,amssymb,amsfonts}
\usepackage{nicefrac}
\usepackage{microtype}
\usepackage[table]{xcolor}
\usepackage{graphicx}
\usepackage{tikz}
\usepackage{textcomp}
\usepackage{algorithm}
\usepackage{algpseudocode}
\usepackage{stfloats}
\usepackage{wrapfig}
\usepackage{enumitem}
\usepackage{longtable}
\usepackage{caption}
\usepackage{subcaption}

\definecolor{ourskyblue}{RGB}{198, 219, 239}

\title{\textbf{\underline{P}rompt \underline{D}riven \underline{E}xploration}\\
{\mdseries Language as an Exploration Space for VLA Reinforcement Learning}}

\author{Sunshine Jiang\textsuperscript{1,3} \quad John Marangola\textsuperscript{1,3} \quad David Zhang\textsuperscript{1,3} \\
Raghuram Kowdeed\textsuperscript{3} \quad Ruiyang Luo\textsuperscript{1,3} \quad Nitish Dashora\textsuperscript{1,3} \\
Richard Li\textsuperscript{1,3} \quad Pulkit Agrawal\textsuperscript{1,3} \quad Zhang-Wei Hong\textsuperscript{1,2,3} \\
\textsuperscript{1}Massachusetts Institute of Technology \quad \textsuperscript{2}MIT-IBM Computing Research Lab \\
\textsuperscript{3}Improbable AI Lab}

\begin{document}\raggedbottom

\maketitle

\begin{abstract}

Exploration is essential to reinforcement learning (RL) since a policy cannot improve by repeatedly sampling the behaviors it already prefers. Standard methods inject stochasticity in the action space, but such jitter only yields rollouts close to the original action distribution. Escaping a weak policy often requires global perturbations that action noise cannot produce. Large language models (LLMs) and vision-language-action (VLA) models offer a pathway: they condition the policy on a natural language prompt, and since the rollout follows from it, modifying the prompt induces global changes. The challenge is finding prompts that induce useful global changes. With a weak policy that rarely succeeds, reward is too sparse to select on. Our idea is to refine prompts from the rollouts themselves: a vision-language model (VLM) reasons over the rollout video, diagnoses how the policy responded, and rewrites the prompt to elicit better behavior next time. Sampling a prompt and holding it fixed throughout a rollout provides episode-level commitment, analogous to posterior-sampling exploration. We call this strategy Prompt Driven Exploration (PDE). Across manipulation and reasoning tasks, PDE enables RL to learn successful policies even from zero-reward starts, and improves sample efficiency more broadly. Our website is available at \url{https://xinyunsunshine.github.io/prompt-rl}.

\end{abstract}

\section{Introduction}

Reinforcement learning (RL) \citep{kaelbling1996reinforcement} has become a dominant post-training paradigm for foundation models because it enables scalable self-improvement beyond supervised learning. RL fine-tuning has unlocked strong reasoning and mathematical capabilities in large language models (LLMs)~\citep{jaech2024openai, Guo2025DeepSeekR1} and enabled direct alignment with human preferences in large diffusion-based text-to-image models~\citep{black2024training}. However, self-improvement is bottlenecked by exploration: a policy can only surpass its current behavior by producing rollouts different from those it already favors, so that RL can reinforce the higher-reward ones.

\begin{figure}[!t]
\centering\vspace{-5pt}
\captionsetup{skip=2pt}\captionsetup[subfigure]{skip=4pt}
\newsavebox{\introConceptBox}\sbox{\introConceptBox}{\includegraphics[width=0.34\textwidth]{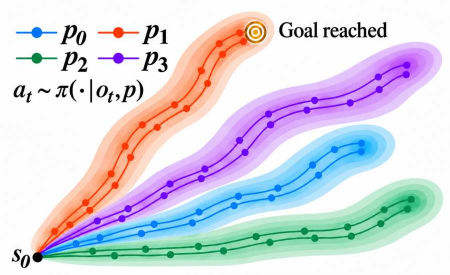}}
\newlength{\introH}\setlength{\introH}{\dimexpr\ht\introConceptBox+\dp\introConceptBox\relax}
\newlength{\introW}\setlength{\introW}{0.655\textwidth}
\begin{subfigure}[b]{0.34\textwidth}
\centering\usebox{\introConceptBox}
\caption{}\label{fig:teaser_global_vs_local}
\end{subfigure}\hspace{0.005\textwidth}%
\begin{subfigure}[b]{0.655\textwidth}
\centering
\begin{tikzpicture}[x=\introW,y=\introH]
\path[use as bounding box] (0,0) rectangle (1,1);
\node[anchor=north west,inner sep=0pt,outer sep=0pt] at (0,1) {\includegraphics[page=1,width=.40\introW,height=.49\introH,keepaspectratio]{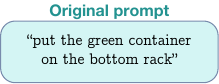}};
\node[anchor=south west,inner sep=0pt,outer sep=0pt] at (0,0) {\includegraphics[page=2,width=.40\introW,height=.49\introH,keepaspectratio]{figure1-prompt-labels.pdf}};
\node[anchor=east,inner sep=0pt,outer sep=0pt] at (1,.72) {\includegraphics[trim=495bp 291bp 728bp 60bp,clip,width=.585\introW]{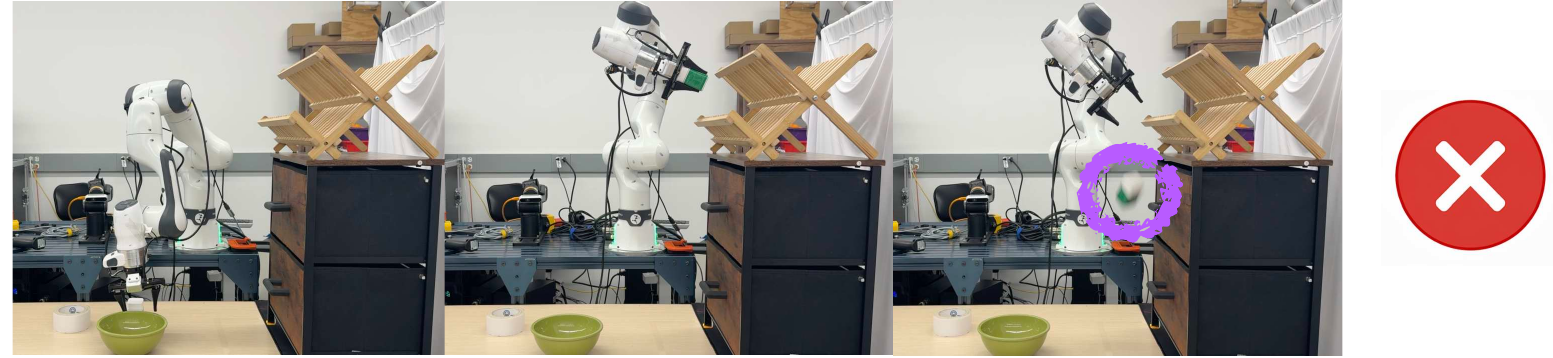}};
\node[anchor=east,inner sep=0pt,outer sep=0pt] at (1,.22) {\includegraphics[trim=495bp 46bp 728bp 305bp,clip,width=.585\introW]{figures/teaser-realworld-archive.pdf}};
\end{tikzpicture}
\caption{}\label{fig:teaser-realworld-example}
\end{subfigure}
\caption{\textbf{Prompt changes elicit different behaviors from the same policy.} (a) Nearby trajectories of the same color show action-noise rollouts under a fixed prompt; different colors show modes induced by different prompts. (b) Adding ``completely'' to the placement instruction turns failure into success without updating policy weights.}

\label{fig:teaser}
\vspace{-6pt}
\end{figure}

To explore beyond the policy's preferred behavior, a standard approach is to sample actions stochastically rather than greedily~\citep{williams1992simple, schulman2017ppo, haarnoja2018soft}. However, such action noise often preserves the overall strategy, or \emph{behavior mode}, of a rollout~\citep{osband2016deep, ecoffet2021first}. Figure~\ref{fig:teaser_global_vs_local} illustrates this \emph{local exploration}, with nearby trajectories within each colored mode representing variations induced by action noise. Success may instead require a different strategy, illustrated by switching from the blue mode to the red mode that reaches the goal. Such \emph{global exploration} requires coordinated changes across the rollout that action noise alone is unlikely to produce. When the initial policy rarely succeeds, local exploration often repeats the same failures, providing little reward signal for improvement.

If action noise primarily produces local exploration, how can we perturb the policy globally? Foundation models offer a pathway. LLMs~\citep{brown2020language, achiam2023gpt} and vision-language-action (VLA) models~\citep{kim2024openvla, black2024pi0} condition their token or action distributions on a natural language prompt. Changing the prompt changes the action distributions across the rollout and can shift exploration from one colored mode to another (Figure~\ref{fig:teaser_global_vs_local}). Figure~\ref{fig:teaser-realworld-example} illustrates how a wording change can elicit useful behavior from a fixed VLA policy. Given the prompt ``put the green container on the bottom rack,'' the policy picks up the container but fails to place it fully on the rack. Changing the prompt to ``put the green container completely on the bottom rack'' leads the policy to place the container fully on the rack and succeed, without any weight updates.  

How can we find prompts that elicit successful behavior? A straightforward approach is to rank candidate prompts by rollout reward and keep the best. When all candidates receive zero reward, this ranking gives no basis for choosing among them. The failed rollout in Figure~\ref{fig:teaser-realworld-example} shows that the policy grasps the container but does not place it fully on the rack, suggesting that the prompt should emphasize completing the placement. Prior work on prompt optimization shows that LLMs can analyze failed outputs and reason over multi-step execution traces to diagnose errors and propose better prompts~\citep{pryzant2023protegi, agrawal2026gepa}.

We propose \textbf{P}rompt \textbf{D}riven \textbf{E}xploration (PDE) to bootstrap RL using feedback from policy rollouts. A vision-language model (VLM) examines where the policy fails and translates the task into prompts that the VLA can act on more effectively. A prompt can help the VLA complete the original task even if it follows the prompt only loosely. If a new prompt elicits even occasional success, RL can reinforce the successful behavior so the policy performs it more often. PDE mirrors the exploration principle of posterior sampling, a well-established exploration algorithm~\citep{strens2000bayesian, osband2013posterior, daniel2018tutorial}. Both explore across rollout distributions. Posterior sampling draws environment models from the posterior and follows their optimal policies; PDE samples prompts that induce different policies. The VLM plays an analogous role to a posterior update, using observed rollouts to refine its prompt proposals~\citep{pryzant2023protegi, agrawal2026gepa}. This guides exploration across prompt-conditioned policies without explicit Bayesian inference.

Our contribution is a simple exploration strategy that enables RL to escape weak initial policies by refining prompts from policy rollouts. We evaluate PDE on LIBERO-PRO~\citep{zhou2025liberopro} and ManiSkill~\citep{tao2024maniskill3}, with an illustrative LIBERO-90 task~\citep{liu2023libero}. PDE enables learning even on tasks where the initial policy has zero success. Across tasks of varying difficulty, PDE achieves higher success rates with far fewer environment interactions and solves tasks where exploration with action noise fails within the training budget. On three real-world manipulation tasks, PDE triples the initial policy's average success rate in under an hour and surpasses the final success rate of PPO with action noise using half as many rollouts (Figure~\ref{fig:realworld-learning-curves}). Our analysis shows that VLM-guided prompt updates reliably shift the prompt distribution toward prompts that elicit success. PDE also accelerates RL on challenging LLM coding tasks, demonstrating that its benefits extend beyond VLA control.

\section{Related Work}

\textbf{RL post-training for VLAs.} Fine-tuning pretrained VLAs under sparse binary rewards is challenging at scale~\citep{tan2025interactive, liu2026what, li2026simplevlarl}. Recent work reduces what RL must update by freezing the backbone and training a small module on top, such as an RL token with an actor-critic head~\citep{xu2025rlt} or residual off-policy actors whose trajectories are distilled back into the base policy~\citep{xiao2026selfimproving}. Other work optimizes in the latent space of a pretrained diffusion policy~\citep{wagenmaker2025steering}. PDE complements them by exploring through prompts.

\textbf{Foundation reward models.} A complementary line provides dense feedback through image- or language-conditioned value functions~\citep{ma2022vip, ma2023liv}, in-context progress estimation~\citep{ma2024vision}, progress supervision on demonstrations~\citep{zhai2025vision}, preference learning~\citep{wang2024rl}, or combined progress and preference supervision~\citep{liang2026robometer}. These methods change the learning signal; PDE changes prompt conditioning while retaining the task reward, so combining them can improve exploration and enrich feedback.

\textbf{Prompt optimization.} LLMs can serve as prompt optimizers that iteratively refine prompts~\citep{yang2023large, zhou2022large}, and analogous methods exist for VLMs through visually grounded discrete optimization~\citep{du2024ipo} and continuous context embeddings~\citep{zhou2022learning}. Closest to our setting, CoVer-VLA~\citep{kwok2026scalingverificationeffectivescaling} uses a trained verifier to select rephrased instructions and action chunks at deployment, and TTT-VLA~\citep{zhang2026ttt} optimizes the latent prompt using a self-supervised learning objective at test time. Prior work jointly optimizes prompts and LLM weights to improve the resulting system~\citep{soylu2024fine}. PDE uses prompts to guide RL exploration, seeking behaviors the policy can learn from rather than an optimal prompt--model pair.

\textbf{RL exploration strategies.} Classical exploration strategies include count-based bonuses~\citep{bellemare2016unifying}, curiosity-driven intrinsic motivation~\citep{pathak2017curiosity}, entropy regularization~\citep{haarnoja2018soft}, and parameter-space noise~\citep{plappert2018parameter}. Parameter-space noise also promotes temporally consistent exploration. PDE explores alternative prompts for the same task, leveraging the language conditioning already present in pretrained VLAs.

\textbf{LLM-guided exploration in RL.}
ExploRLLM \citep{ma2025explorllm} uses LLM-generated policy code to guide exploration, while LLM-TALE \citep{luijkx2026llmtale} generates task-level and affordance-level plans that are refined online. SENSEI \citep{sancaktar2025sensei} distills VLM judgments of interestingness into an intrinsic reward, whereas PR2L \citep{chen2025pr2l} and ConPE \citep{choi2023conpe} use prompts to construct representations for downstream policy learning and adaptation. VLGOR \citep{liu2026vlgor} instead generates imagined rollouts to augment offline RL. Prior work~\citep{arumugam2026toward} uses an LLM as the decision-making agent, updating beliefs about the environment to guide action selection. PDE instead uses a VLM to propose prompts that guide a separate policy's exploration, then uses RL on the resulting rollouts to reinforce successful behavior in that policy.

\section{Preliminaries}
\label{sec:prelim}

We fine-tune a vision-language-action (VLA) policy $\pi_\theta$~\citep{black2024pi0,kim2024openvla} with reinforcement learning (RL)~\citep{kaelbling1996reinforcement} in a multi-task setting. Each task $g$ is paired with a canonical prompt $p_g$. At the start of each rollout, we sample a task $g \sim p_{\mathcal{G}}$, where $p_{\mathcal{G}}$ is the task distribution, and provide its task prompt $p_g$ to the policy. We write $\pi_\theta(\cdot \mid p_g)$ for the policy conditioned on this prompt. At each timestep $t$, the policy receives an observation $o_t$, such as RGB images or text, and outputs an action $a_t$. We denote the resulting rollout by
$\tau = (o_0, a_0, \dots, o_{T-1}, a_{T-1}, o_T)$.
The policy receives a terminal reward
\[
R(\tau, g) =
\begin{cases}
1, & \text{if } \tau \text{ completes task } g,\\
0, & \text{otherwise}.
\end{cases}
\]

\textbf{Reward depends on the task, not the prompt.}
For a fixed rollout, the reward depends on the task $g$, not on the prompt used to generate it. For example, if $g$ is ``open the drawer,'' any rollout that opens the drawer receives reward $1$, whether the prompt is the canonical prompt $p_g = $ ``open the drawer'' or an alternative such as ``pull the handle toward you.'' Thus, a prompt need not faithfully describe $g$ to be useful. It only needs to elicit a rollout that completes $g$. Section~\ref{sec:method} shows how our method exploits this flexibility for exploration.
{
The RL objective is to maximize expected success:
\begin{align}
    J(\pi_\theta) = \mathbb{E}_{\,g \sim p_{\mathcal{G}},\; \tau \sim \pi_\theta(\cdot \mid p_g)}\!\left[\, R(\tau, g) \,\right].
\end{align}

}

\section{Method: Prompt Driven Exploration (PDE)}
\label{sec:method}

\textbf{Goal.} Under the canonical prompt $p_g$, a policy may rarely sample trajectories that complete task $g$. Prompt Driven Exploration (PDE) varies the prompt while keeping the task and its reward fixed. The challenge is to find prompts that elicit successful behavior when the policy rarely succeeds under $p_g$.

\begin{figure}[tb]
    \centering
    \begin{tikzpicture}
    \node[anchor=south west,inner sep=0] (pdeimage) at (0,0)
        {\includegraphics[width=\textwidth]{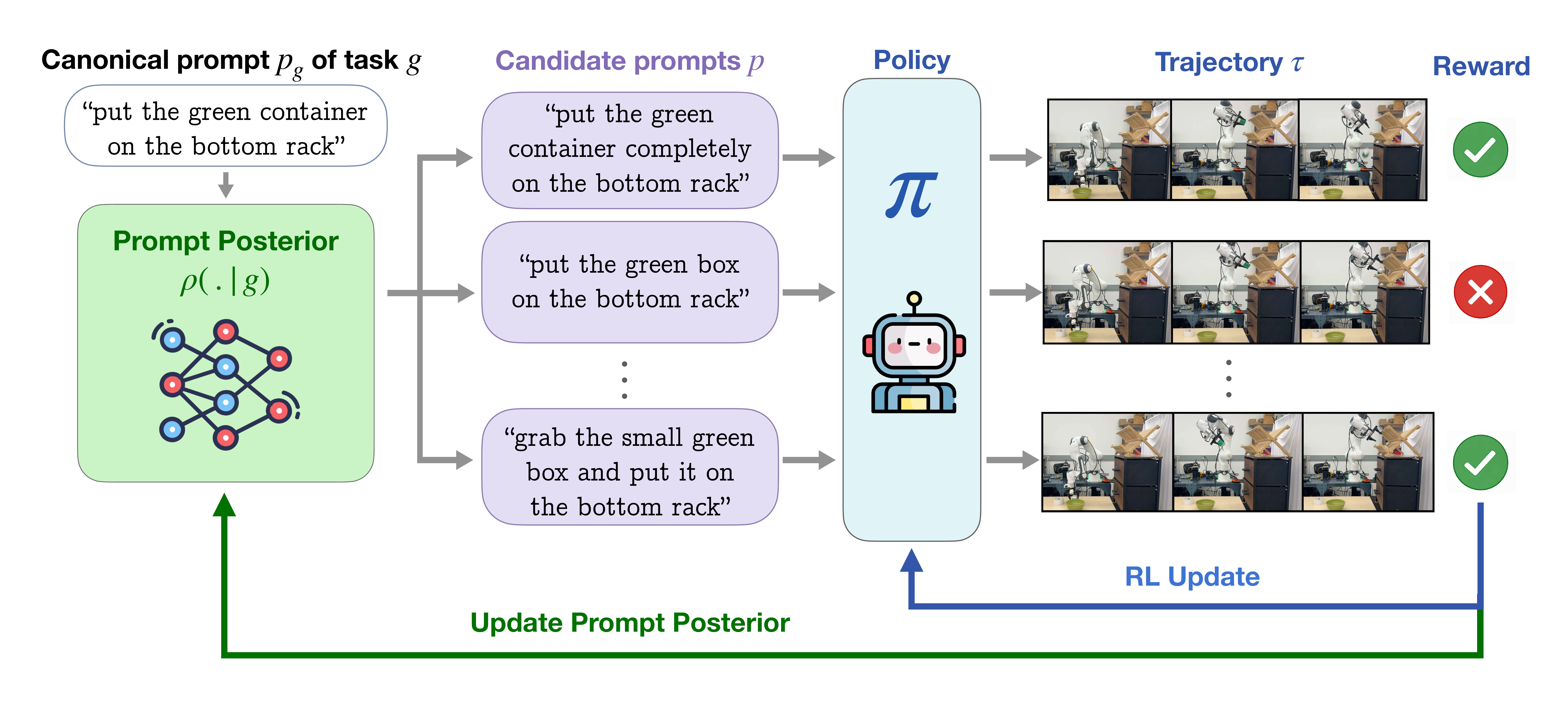}};
    \begin{scope}[x={(pdeimage.south east)},y={(pdeimage.north west)}]
        \fill[fill={rgb,255:red,184;green,238;blue,189}] (0.055,0.555) rectangle (0.235,0.691);
        \node[font=\sffamily\bfseries\fontsize{7}{8}\selectfont,text=green!50!black]
            at (0.145,0.658) {Prompt Sampler};
        \node[font=\fontsize{6}{7}\selectfont,text=teal!55!black]
            at (0.145,0.594) {$\rho(\cdot\mid g)$};
        \fill[white] (0.291,0.088) rectangle (0.512,0.150);
        \node[font=\sffamily\bfseries\fontsize{7}{8}\selectfont,text=green!50!black]
            at (0.402,0.119) {Prompt Refinement};
    \end{scope}
    \end{tikzpicture}
    \caption{\textbf{Overview.} The VLM serves as the prompt sampler $\rho(\cdot \mid g)$ and refines its proposals using the canonical prompt $p_g$ and previous rollout feedback. The VLA policy $\pi_\theta$ executes sampled prompts, and the resulting trajectories and rewards inform subsequent prompt proposals (\textcolor{green!50!black}{green arrow}). RL updates the policy using the collected rollouts (\textcolor{blue}{blue arrow}).}
    \label{fig:pipeline}
\end{figure}

\textbf{Our approach.}
As illustrated in Fig.~\ref{fig:pipeline}, PDE uses a VLM to refine a prompt sampler $\rho_i(\cdot\mid g)$ from rollout feedback. Starting from rollouts under the canonical prompt $p_g$, the VLM diagnoses where the policy fails and proposes prompts that may help it complete $g$ (\textbf{Prompt refinement}, green arrow). The VLM thus acts as a \textit{translator}, using the policy's rollouts to express the task in a form that elicits the intended behavior. Reward depends on completing the original task, not on following the prompt exactly. Even prompts with grammatical errors or unusual meanings can be useful if they elicit successful behavior. Better prompts may elicit success only occasionally. RL updates $\pi_\theta$ on the collected rollouts to reinforce these successes and make the behavior more reliable (\textbf{RL Update}, blue arrow). The updated policy's rollouts and rewards guide further prompt refinement (Algorithm~\ref{alg:pde}).

\subsection{Exploration in the Prompt Space}
\label{sec:framework}

PDE first refines the prompt sampler $\rho(\cdot\mid g)$ with the policy held fixed to bootstrap exploration. It can continue refining the sampler as the policy learns. In both stages, the VLM uses the target task $g$ and past rollouts to propose prompts that address observed failures. We store each task, prompt, trajectory, and reward as a tuple $(g,p,\tau,R(\tau,g))$ in the interaction history $\mathcal{H}$.

\textbf{Bootstrapping.}
PDE collects rollouts under the canonical prompt $p_g$ and adds their trajectories and rewards to $\mathcal{H}$. The VLM examines these rollouts to identify how the policy fails and proposes alternative prompts that may address those failures. During bootstrapping, we keep $\theta$ fixed and evaluate the proposed prompts by collecting new rollouts. These trajectories and rewards are added to $\mathcal{H}$ so the VLM can assess how its proposals changed the policy's behavior and refine the sampler. Policy optimization begins after this bootstrapping stage.

\textbf{Prompt refinement.}
After bootstrapping, PDE adapts prompt sampling to the policy's success under the canonical prompt $p_g$. We call prompts sampled from the VLM-refined sampler $\rho(\cdot\mid g)$ \emph{exploratory prompts}. When success under $p_g$ is rare, these prompts can expose successful behavior for RL to reinforce. At test time, the policy receives only the canonical prompt $p_g$, so success under exploratory prompts alone is insufficient. As the success rate under $p_g$ increases, PDE collects more rollouts under this prompt to make successful behavior more reliable. We sample from the mixture
\begin{align}
p \sim \alpha \delta_{p_g} + (1-\alpha)\rho(\cdot \mid g),\label{eq:prompt-mixture}
\end{align}
where $\alpha \in [0,1]$ is the probability of using $p_g$. We set $\alpha$ as an increasing function of the success rate under $p_g$ (Appendix~A.3). This balances exploration through alternative prompts with learning under the canonical prompt. Each sampled prompt is held fixed throughout its rollout, and the resulting tuple $(g,p,\tau,R(\tau,g))$ is added to $\mathcal{H}$.

The VLM can also refine $\rho$ during policy optimization, using behavioral summaries of rollout frames and task success rates to diagnose failures and propose prompt revisions~\citep{shinn2023reflexion,yuksekgonul2024textgrad}. These summaries help the VLM distinguish failure modes even when rollouts receive the same reward. Appendix~\ref{app:prompts} gives the prompt templates. In our VLA experiments, we keep $\rho$ fixed after bootstrapping while continuing to adapt $\alpha$ (Appendix~\ref{app:schedule}).

\subsection{Combining PDE with Policy Optimization}
\label{sec:mixed}

We implement the RL update in Algorithm~\ref{alg:pde} with Proximal Policy Optimization (PPO)~\citep{schulman2017ppo}. For a rollout $\tau=\{(o_t,a_t)\}_{t=0}^{T-1}$ collected under prompt $p$, PPO minimizes
\begin{equation}
\begin{aligned}
\mathcal{L}^{\mathrm{PPO}}(\theta)
&=
-\mathbb{E}_{t}\!\left[
\min\!\left(
w_t(\theta)\hat{A}_t,
\operatorname{clip}(w_t(\theta),1-\epsilon,1+\epsilon)\hat{A}_t
\right)
\right], 
w_t(\theta)
=
\frac{\pi_\theta(a_t\mid o_t,p)}
{\pi_{\mathrm{old}}(a_t\mid o_t,p)} ,
\end{aligned}
\end{equation}
where $\hat{A}_t$ is the advantage estimate, $\epsilon$ the clipping threshold, and $\pi_{\mathrm{old}}$ the rollout policy.

\textbf{Mixed backpropagation.}
A standard PPO update on a rollout collected under an exploratory prompt $p$ directly trains $\pi_\theta(\cdot\mid o,p)$, but may not improve $\pi_\theta(\cdot\mid o,p_g)$. To update the policy under both prompts from the same rollout, we average their current-policy log-probabilities and use this average in the PPO ratio,
\begin{equation}
\begin{aligned}
\ell_t(\theta)
&=
\frac{1}{2}\log\pi_\theta(a_t\mid o_t,p_g)
+
\frac{1}{2}\log\pi_\theta(a_t\mid o_t,p), \\
w_t(\theta)
&=
\exp\!\left(
\ell_t(\theta)
-
\log\pi_{\mathrm{old}}(a_t\mid o_t,p)
\right).
\end{aligned}
\end{equation}
The denominator uses the probability recorded under the rollout prompt $p$. Using the geometric mean of action probabilities gives equal weight to the log-probability gradients under both prompts. An arithmetic mean would instead favor the prompt that already assigns higher probability to the sampled action. Appendix~\ref{app:mixed-backprop} analyzes this weighting and its implications for policy optimization.

\subsection{Connection to Posterior Sampling}
\label{sec:posterior-sampling}
Posterior sampling for reinforcement learning (PSRL) explores by sampling policies from the distribution induced by a posterior over environment models~\citep{strens2000bayesian,osband2013posterior}. PDE implements a similar exploration principle through the prompt distribution $\rho(\cdot\mid g)$. Each prompt $p$ specifies a policy $\pi_\theta(\cdot\mid o,p)$ and its associated rollout distribution. Sampling a prompt thus selects a rollout distribution, paralleling policy selection in PSRL. Prior work demonstrates that LLMs can approximate posterior updates from interaction history~\citep{arumugam2026toward}. The VLM serves an analogous function in PDE, using rollout evidence to refine the prompt distribution and guide subsequent exploration across policies.

\begin{algorithm}[t]
\caption{Prompt Driven Exploration (PDE).}
\label{alg:pde}
\begin{algorithmic}[1]
\Require Policy $\pi_\theta$, task distribution $p_{\mathcal{G}}$, VLM
\State Bootstrap prompt sampler $\rho$ and rollout history $\mathcal{H}$ with $\theta$ fixed (Section~\ref{sec:framework})
\While{training budget remains}
  \State Sample task $g\sim p_{\mathcal{G}}$ and prompt $p$ using the mixture in Eq.~\eqref{eq:prompt-mixture}
  \State Collect rollout $\tau$ under $p$ and evaluate $R(\tau,g)$
  \State Add $(g,p,\tau,R(\tau,g))$ to $\mathcal{H}$
  \State When enabled, refine $\rho$ with the VLM using $\mathcal{H}$
  \State Update $\theta$ on $\tau$ (Section~\ref{sec:mixed})
\EndWhile
\State \Return $\pi_\theta$
\end{algorithmic}
\end{algorithm}

\section{Experiments}
\label{sec:experiments}

{
We evaluate PDE through three questions: (i) Can changing the prompts elicit successful behavior when the initial policy fails under the canonical prompt? (ii) Does PDE improve RL across tasks and base models? (iii) Do the gains transfer to the canonical prompt used at deployment? We also ablate key design choices (Section~\ref{subsec:analysis}) and validate PDE on real-world manipulation tasks (Section~\ref{subsec:exp:real}).
}

\subsection{Setup}
\label{sec:exp-setup}

\textbf{Environment and Metric.}
We evaluate on LIBERO-PRO~\citep{zhou2025liberopro} and ManiSkill~\citep{tao2024maniskill3}, with an illustrative case study on a LIBERO-90 task~\citep{liu2023libero}. We report episode-level success using the RLinf evaluation suite\footnote{\url{https://rlinf.readthedocs.io/}}: an episode succeeds if it completes the task at least once (\texttt{success\_once}). Evaluation counts are provided in Appendix~\ref{app:exp}. The LIBERO-PRO learning curves report training success; policy evaluation uses only the canonical task prompt $p_g$. We start RL training with an initial policy from a weak $\pi_{0.5}$ SFT checkpoint~\citep{chen2025pirl}
trained with a limited number of demonstrations from four LIBERO task suites, excluding LIBERO-90, unless specified.

\textbf{Baselines.}
For each backbone, PDE and the baselines start from the same SFT checkpoint and use the same PPO~\citep{schulman2017ppo} optimizer settings and policy training rollout budget. The baselines train under the canonical prompt $p_g$. PDE additionally collects bootstrapping rollouts; these are included in the learning curves, and their cost is reported separately in Appendix~\ref{app:exp:compute}. Checkpoint and implementation details are provided in the corresponding result sections and Appendix~\ref{app:exp}; hyperparameters are listed in Appendix~\ref{app:hparams}.

\begin{itemize}[leftmargin=*, itemsep=1pt, topsep=2pt]
    \item \textbf{Noise}: \textbf{Action Noise} is standard PPO with stochastic action sampling. \textbf{Parameter Noise} applies Gaussian perturbations~\citep{plappert2018parameter} to the policy weights at the beginning of each rollout.
    \item \textbf{Bonus}: Random Network Distillation (RND)~\citep{burda2018exploration} adds intrinsic reward, computed as the prediction error of a learned predictor against a fixed random target network.
    \item \textbf{Dense reward}: \textbf{Robometer}~\citep{liang2026robometer} provides dense task-progress rewards from egocentric images and the task description.
\end{itemize}

\subsection{Illustrative Experiment: Close the Microwave}
\label{sec:exp-microwave}

\begin{figure}[!htbp]
    \centering
    \includegraphics[height=0.26\textwidth,trim=0 0 1291 0,clip]{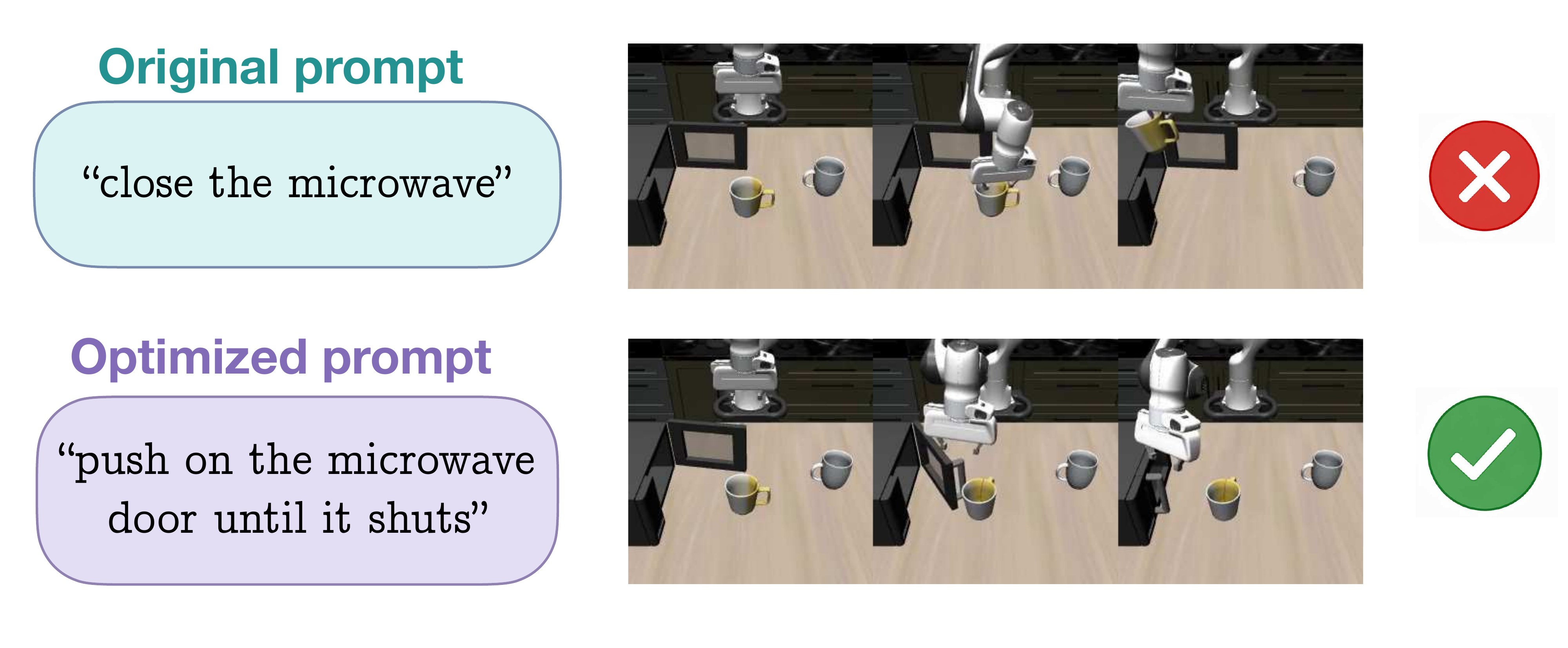}\hspace{0.015\textwidth}%
    \includegraphics[height=0.26\textwidth]{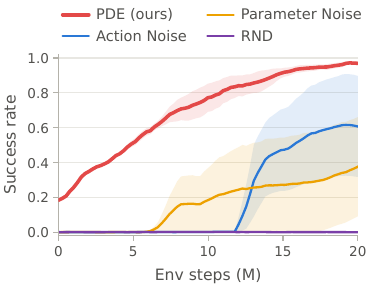}
    \caption{\textbf{Left:} Under ``close the microwave,'' the policy grasps the mug (top). PDE discovers ``push on the microwave door until it shuts,'' eliciting successful door closing without weight updates (bottom). \textbf{Right:} PDE accelerates learning and achieves higher training success than action noise as the sampling schedule shifts toward the canonical prompt.}
    \label{fig:teaser-illustrative-microwave}
\end{figure}

\textbf{Setup.}
{
We first examine a LIBERO-90 task with a clear failure mode: $g=$ \emph{``close the microwave''}, with canonical prompt $p_g=g$ (Figure~\ref{fig:teaser-illustrative-microwave}). The initial $\pi_{0.5}$ policy described in Section~\ref{sec:exp-setup} has no observed success under this prompt. PPO with action noise eventually improves, but makes little progress early in training.
}

{
\textbf{How does the initial policy fail?}
Under the canonical prompt, the arm grasps the yellow-handled mug, moves it toward the microwave, and stalls without contacting the door (Figure~\ref{fig:teaser-illustrative-microwave}, top). This behavior resembles the visually similar training task \emph{``put the yellow and white mug in the microwave and close it.''} This suggests that the policy defaults to a familiar pick-and-place behavior rather than attempting the door-closing interaction.
}

{
\textbf{Can PDE discover successful prompts?}
During bootstrapping, PDE refines prompts while keeping the initial policy fixed. For example, ``push on the microwave door until it shuts'' elicits door pushing instead of mug grasping, producing successful rollouts without weight updates (Figure~\ref{fig:teaser-illustrative-microwave}, bottom). PDE finds multiple successful prompts across independent runs (Figure~\ref{fig:prompt_discovery}). The change from mug grasping to door pushing illustrates a shift in the overall behavior attempted, without changing the policy weights or task reward. The successful prompts emphasize three types of information:
}
\begin{itemize}[leftmargin=*, itemsep=1pt, topsep=2pt]
    \item \textbf{Explicit contact and action}: prompts such as
    \emph{``push the microwave door closed,''} \emph{``push the microwave door until it
    clicks,''} and \emph{``swing the microwave door shut''} specify the desired physical
    interaction, shifting the policy from pick-and-place behavior to
    push-and-close behavior.

    \item \textbf{Spatial or generic reference}: prompts such as
    \emph{``close the black appliance door on the left,''} \emph{``close the kitchen
    appliance door,''} and \emph{``shut the open appliance door''} identify the target
    by appearance or location rather than by the object name \emph{``microwave,''}
    providing an alternative description of the same target.

    \item \textbf{Distractor exclusion}: prompts such as
    \emph{``do not close the cabinet -- close microwave''} and \emph{``first move to the
    microwave, then close the door''} explicitly redirect the policy away from
    distractors and toward the intended contact point.
\end{itemize}

\textbf{Can PDE bootstrap RL from zero success?} The discovered prompts yield successful rollouts, but do not yet make the frozen policy reliable. PPO with PDE builds on these early successes and reaches near-perfect training success, while Action Noise improves later and plateaus lower (Figure~\ref{fig:teaser-illustrative-microwave}, right).

\subsection{Does PDE improve learning across task difficulties and VLA models?}
\label{sec:exp-liberopro}

\begin{figure}[!htbp]
      \centering
      \includegraphics[width=\textwidth]{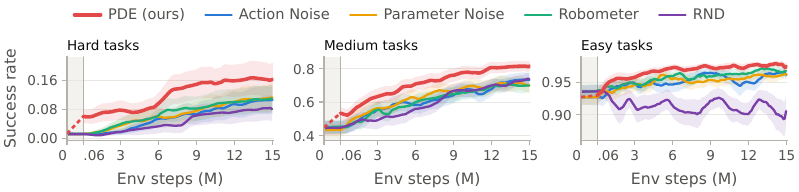}
    \caption{\textbf{$\pi_{0.5}$ learning curves on LIBERO-PRO.} PDE improves learning across task difficulties, with the largest relative gain over action noise on hard tasks. Curves show mean training success across tasks; shading indicates standard error. Panels show hard, medium, and easy tasks from left to right. Dashed red segments mark the $0.06$M bootstrapping steps, with this interval stretched for visibility.}                                 \label{fig:agg_task_difficulty}
  \end{figure}

\begin{figure}[!t]
    \centering
    \begin{subfigure}{\textwidth}
        \centering
        \includegraphics[width=\linewidth]{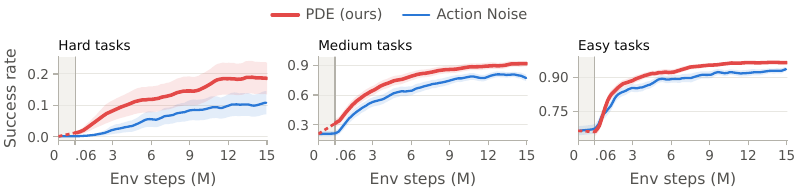}
        \captionsetup{skip=2pt}
        \caption{GR00T}
        \label{fig:agg_task_difficulty_gr00t}
    \end{subfigure}
    \par\vspace{-3pt}
    \begin{subfigure}{\textwidth}
        \centering
        \begin{tikzpicture}\node[anchor=south west,inner sep=0] (pirow) at (0,0) {\includegraphics[width=\linewidth]{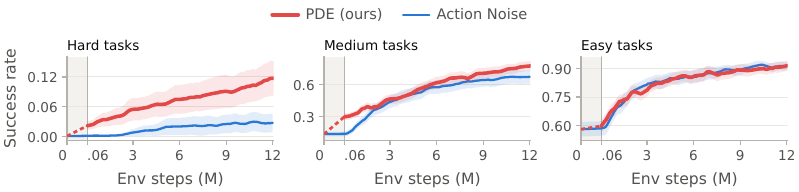}}; \begin{scope}[x={(pirow.south east)},y={(pirow.north west)}]\fill[white] (0,0.88) rectangle (1,1);\end{scope}\end{tikzpicture}
        \captionsetup{skip=2pt}
        \caption{$\pi_0$}
        \label{fig:agg_task_difficulty_pi0}
    \end{subfigure}
    \caption{\textbf{PDE improves learning across VLA backbones.} PDE improves over action noise on GR00T and $\pi_0$, with clear gains on hard and medium tasks and smaller gains on easy tasks. Rows show GR00T (top) and $\pi_0$ (bottom); columns show hard, medium, and easy tasks. Curves report mean training success across tasks, with shading denoting standard error. Dashed red segments mark the $0.06$M bootstrapping steps, with this interval stretched for visibility.}
    \label{fig:agg_task_difficulty_backbones}
\end{figure}

\textbf{Setup.}
We evaluate adaptation on LIBERO-PRO~\citep{zhou2025liberopro} using object, swap, and task perturbations, training a separate multitask policy for each benchmark group. We group tasks as hard, medium, or easy by each model's initial success rate to assess how PDE's benefit varies with initial performance. Appendix~\ref{app:exp} provides training details.

\textbf{How does PDE compare with other exploration methods?} Figure~\ref{fig:agg_task_difficulty} shows the largest relative gain over Action Noise on hard tasks, where the initial policy rarely succeeds. The results with the policy held fixed during bootstrapping (Table~\ref{tab:prompt-only-ceiling}) show that prompt refinement already elicits successful rollouts before RL begins, providing a learning signal when action noise yields few successes. On medium tasks, PDE accelerates learning even when Action Noise makes progress; on easy tasks, high initial success leaves less room for improvement. Table~\ref{tab:libero-aggregated-pi05} reports evaluation success rates under the canonical prompt. Parameter Noise holds a weight perturbation fixed throughout each rollout, yet achieves lower mean evaluation success than PDE. This suggests that maintaining a consistent perturbation is less effective here than guiding exploration through prompts. RND encourages novelty, while Robometer supplies dense task feedback, but neither consistently improves evaluation success over Action Noise. Their results suggest that rewarding unfamiliar states or providing richer feedback does not necessarily elicit the behaviors needed for task completion. PDE uses rollout feedback to select prompts that elicit useful behavior, directly addressing which trajectories the policy explores.

\textbf{Does PDE improve learning across base models?} The benefit extends to GR00T and $\pi_0$ (Figure~\ref{fig:agg_task_difficulty_backbones}; Table~\ref{tab:libero-aggregated-gr00t-pi0}). Both models benefit on hard and medium tasks, while the easy-task curves are closer. $\pi_0$'s absolute success on hard tasks remains low despite its advantage over Action Noise. Model-specific update settings are provided in Appendix~\ref{app:anchoring}.

\subsection{Does PDE improve adaptation to different distribution shifts?}
\label{sec:exp-maniskill}

\textbf{Setup.} {We use ManiSkill~\citep{tao2024maniskill3} to test adaptation beyond LIBERO-PRO on a tabletop task that requires placing objects onto a receptacle. Following Liu et al.~\citep{liu2026what}, we evaluate shifts in visual appearance (backgrounds and textures), semantics (objects, instructions, and distractors), and execution (initial states, robot poses, and disturbances). Policies are fine-tuned and evaluated within each shifted setting; Appendix~\ref{app:maniskill-setup} gives the full protocol.}

\textbf{Does PDE improve over Action Noise?}
{We aggregate results by initial task difficulty and type of distribution shift, with difficulty thresholds given in Appendix~\ref{app:difficulty-groups}.} Table~\ref{tab:maniskill-aggregated}
reports success rates (\%) for both views. {PDE outperforms Action Noise in every difficulty tier and type of distribution shift.} Although Action Noise already reaches nontrivial
success on ManiSkill hard tasks, {PDE improves the aggregate success rate across}
the full benchmark. These results show that varying prompts can improve learning
even when action noise already yields successful rollouts.

\begin{table}[!htbp]
  \centering
  \footnotesize
  \setlength{\tabcolsep}{4pt}
  \renewcommand{\arraystretch}{0.95}
    \caption{{\textbf{Adaptation on ManiSkill.} PDE outperforms action noise across difficulty tiers and distribution shifts.}}
  \label{tab:maniskill-aggregated}
  \resizebox{\textwidth}{!}{
  \begin{tabular}{lccc|ccc|c}
  \toprule
  & \multicolumn{3}{c|}{\textbf{Difficulty}} & \multicolumn{3}{c|}{{\textbf{Distribution shift}}} & \\
  \cmidrule{2-4} \cmidrule{5-7}
  & \textbf{Hard} & \textbf{Medium} & \textbf{Easy} & \textbf{Visual} & \textbf{Semantic} & \textbf{Execution} & \textbf{All} \\
  \midrule
  Initial policy & $12.17{\scriptstyle \pm 0.87}$ & $35.16{\scriptstyle \pm 1.08}$ & $64.96{\scriptstyle \pm 1.48}$ & $48.16{\scriptstyle \pm 2.59}$ & $29.58{\scriptstyle \pm 3.17}$ & $28.31{\scriptstyle \pm 2.54}$ & $37.59{\scriptstyle \pm 1.76}$ \\
  \midrule
  Action Noise & $51.65{\scriptstyle \pm 2.56}$ & $63.39{\scriptstyle \pm 2.57}$ & $71.73{\scriptstyle \pm 2.95}$ & $72.12{\scriptstyle \pm 2.24}$ & $54.64{\scriptstyle \pm 3.21}$ & $53.96{\scriptstyle \pm 2.73}$ & $62.31{\scriptstyle \pm 1.67}$ \\
  \rowcolor{ourskyblue}
  PDE (Ours) & $\mathbf{58.77}{\scriptstyle \pm 2.73}$ & $\mathbf{76.68}{\scriptstyle \pm 2.58}$ & $\mathbf{84.33}{\scriptstyle \pm 2.14}$ & $\mathbf{85.12}{\scriptstyle \pm 1.86}$ & $\mathbf{63.65}{\scriptstyle \pm 3.29}$ & $\mathbf{63.75}{\scriptstyle \pm 2.70}$ & $\mathbf{73.32}{\scriptstyle \pm 1.64}$ \\
  \bottomrule
  \end{tabular}
  }
\end{table}

\subsection{Analysis}
\label{subsec:analysis}

{\textbf{Does prompt diversity alone explain the gain?} We replace prompts found during bootstrapping with five random VLM-generated paraphrases per task, without rollout feedback or success filtering. The remaining training procedure is unchanged. Random paraphrases perform similarly to Action Noise, while PDE achieves higher training success (Figure~\ref{fig:random-prompt-baseline}). The gap supports using rollout feedback to find useful prompts rather than relying on varied wording alone.}

{\textbf{Does adaptive sampling help after bootstrapping?} We use the same prompts in both variants but replace adaptive prompt sampling with fixed uniform sampling. Adaptive sampling achieves higher training success (Figure~\ref{fig:random-prompt-baseline}), showing that the sampling schedule matters even after useful prompts have been found.}

{\textbf{Do the gains transfer to the canonical prompt?} Training with the prompts found during bootstrapping and gradually increasing the use of $p_g$ yields higher evaluation success under $p_g$ than training only with the single best prompt (Table~\ref{tab:curriculum_deploy}). This comparison tests the prompt set and sampling schedule together. In a separate ablation, mixed backpropagation achieves higher training success than updates under only the canonical or exploratory prompt (Table~\ref{tab:ablation-object-task}), supporting the use of both prompts in policy optimization.}

\begin{table}[!htbp]
  \centering
  \small
  \setlength{\tabcolsep}{3pt}
  \renewcommand{\arraystretch}{1.15}
    \caption{{\textbf{Illustrative learning behaviors.} Success rates (\%) for three LIBERO-PRO Object tasks with task perturbations. Step 0 is the start of RL, before any policy updates. \textbf{Regime 1:} PDE learns faster than Action Noise, which eventually succeeds. \textbf{Regime 2:} PDE starts with some successes and improves through RL, while Action Noise stays at zero. \textbf{Regime 3:} Both start at zero, but only PDE learns within the training budget. Task details are provided in Appendix~\ref{sec:exp-ablation}.}}
  \label{tab:libero-regime-breakdown}
  \begin{tabular*}{\textwidth}{@{\extracolsep{\fill}}lccc@{\hspace{8pt}}ccc@{\hspace{8pt}}ccc@{}}
    \toprule
    & \multicolumn{3}{c}{\textbf{Regime 1}}
    & \multicolumn{3}{c}{\textbf{Regime 2}}
    & \multicolumn{3}{c}{\textbf{Regime 3}} \\
    \cmidrule(lr){2-4} \cmidrule(lr){5-7} \cmidrule(lr){8-10}
    \textbf{Method}
    & Step 0 & Step 30 & Step 120
    & Step 0 & Step 30 & Step 120
    & Step 0 & Step 30 & Step 120 \\
    \midrule
    Action Noise    & $0.0$ & $65.2$ & $93.5$ & $0.0$ & $0.0$ & $0.0$ & $0.0$ & $0.0$ & $0.0$ \\
    Parameter Noise & $0.0$ & $0.0$ & $0.0$ & $33.3$ & $7.8$ & $0.0$ & $47.1$ & $41.4$ & $0.0$ \\
    Robometer       & $0.0$ & $42.3$ & $97.6$ & $0.0$ & $0.0$ & $0.0$ & $0.0$ & $0.0$ & $0.0$ \\
    RND             & $0.0$ & $0.0$ & $6.2$ & $0.0$ & $0.0$ & $0.0$ & $0.0$ & $0.0$ & $0.0$ \\
    \midrule
    \rowcolor{ourskyblue}
    \textbf{PDE (Ours)}
    & $24.4$ & $\mathbf{99.0}$ & $\mathbf{100.0}$
    & $4.3$ & $\mathbf{29.5}$ & $\mathbf{77.7}$
    & $0.0$ & $0.0$ & $\mathbf{81.8}$ \\
    \bottomrule
  \end{tabular*}
\end{table}

{
Table~\ref{tab:libero-regime-breakdown} illustrates three learning patterns on individual tasks; Appendix~\ref{sec:exp-ablation} gives the task details.
}

{
\textbf{Faster learning when action noise can eventually succeed.}
In Regime~1, both methods eventually attain high success, but PDE reaches it much earlier. Exploratory prompts elicit successful behavior before policy updates begin, giving RL a stronger starting point.
}

{
\textbf{Bootstrapping when action noise fails.}
In Regime~2, PDE starts with occasional successes and improves through RL, while Action Noise, Robometer, and RND show no success at the reported checkpoints. Here, alternative prompts provide the successful rollouts that RL needs to improve.
}

{
{\textbf{Learning despite zero observed early success.}}
In Regime~3, PDE has no observed success at the first two reported checkpoints, but learns by the final checkpoint. Action Noise, Robometer, and RND remain at zero. PDE can therefore learn even when bootstrapping yields no observed success on that task. This result alone does not distinguish transfer across tasks from successful exploration later in training.
}

\subsection{Real-World Manipulation}
\label{subsec:exp:real}

\textbf{Setup.}
{We test whether PDE also improves learning with a small budget of physical robot rollouts, where unsuccessful exploration consumes execution time.}
Our hardware setup consists of a Franka FR3 arm, a set of manipulable tabletop objects, two drawers, and a two-level rack, as shown in Figure~\ref{fig:setup-realworld}. We first perform supervised fine-tuning (SFT) of $\pi_{0.5}$ on 10 language-conditioned manipulation tasks that cover a range of difficulty and horizon lengths, from short-horizon pick-and-place tasks to longer multi-stage tasks involving drawers and spatial constraints. Starting from the SFT checkpoint, we then continue training with either PPO using action noise or PPO augmented with PDE. Additional implementation details are provided in Appendix~\ref{sec:exp-real}.

\begin{figure}[!htbp]
    \centering
    \begin{subfigure}[b]{0.43\textwidth}
        \centering
        \includegraphics[width=\linewidth]{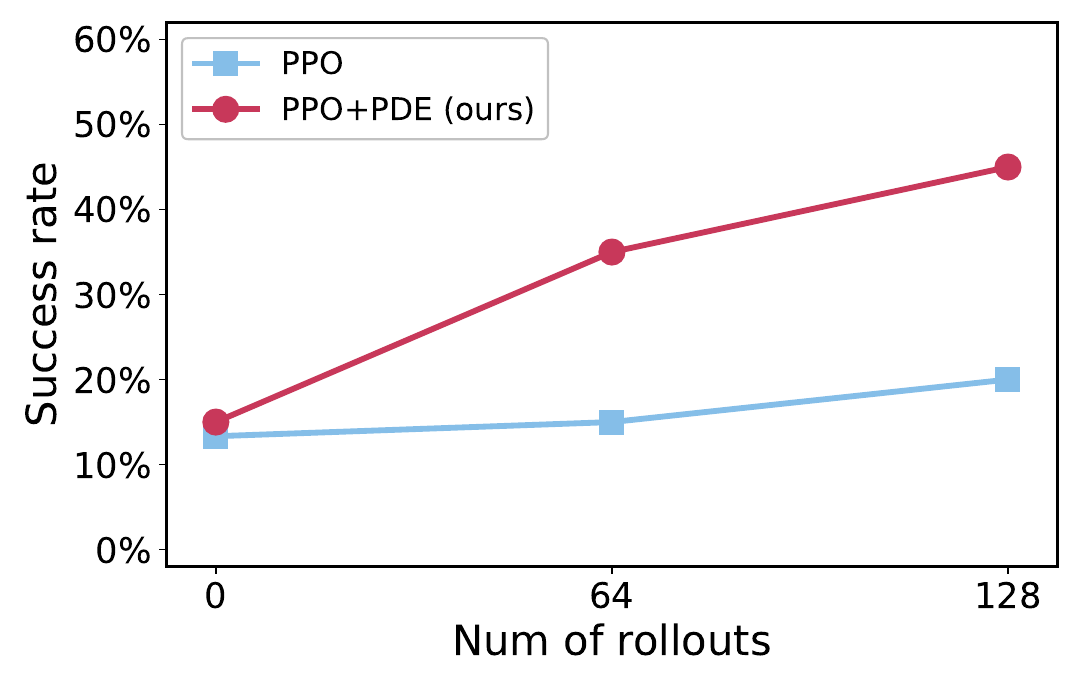}
        \caption{}
        \label{fig:realworld-learning-curves}
    \end{subfigure}
    \hfill
    \begin{subfigure}[b]{0.55\textwidth}
        \centering
                \includegraphics[width=\linewidth]{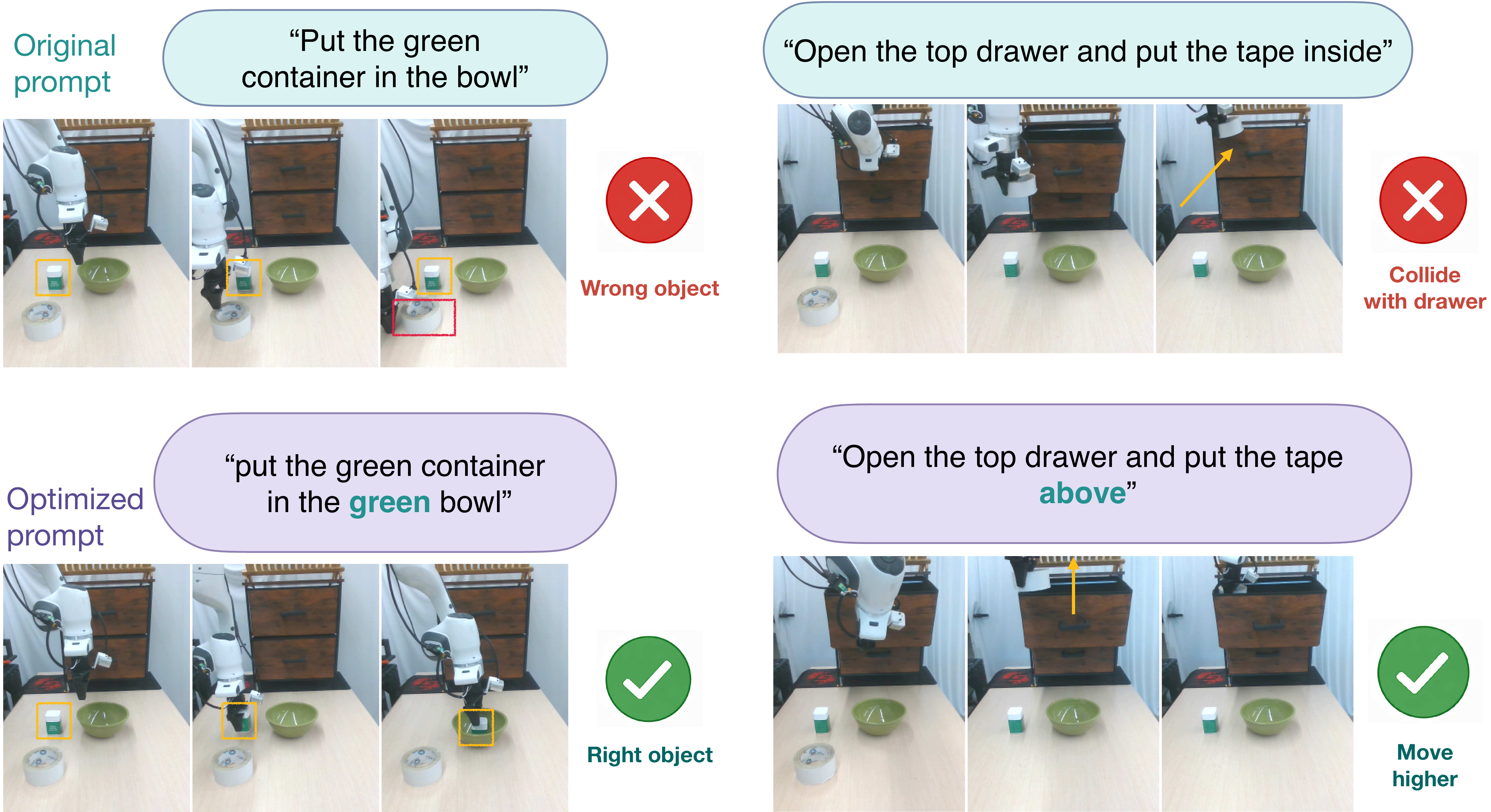}
        \caption{}
        \label{fig:realworld-failures}
    \end{subfigure}
    \caption{\textbf{Real-world manipulation results.} \textbf{(a)} Mean success across real-world tasks: PPO+PDE learns faster and achieves more than twice the success of action noise after 128 rollouts per task. \textbf{(b)} PDE refines object references and spatial instructions to correct failures under the original prompts.}
    \label{fig:realworld}
\end{figure}

{
\textbf{Results.}
PPO with PDE learns faster than PPO with action noise across the three real-world tasks (Figure~\ref{fig:realworld-learning-curves}). After 128 training rollouts per task, its average success rate is more than twice that of the baseline. Final per-task evaluations also show improvement over the SFT policy on every task, whereas action noise provides mixed gains (Appendix~\ref{sec:exp-real}).
}

{
The qualitative examples in Figure~\ref{fig:realworld-failures} illustrate how prompt refinement changes the policy's behavior. For ``put the green container in the bowl,'' PDE adds the bowl's color to make the target reference more specific. For ``open the top drawer and put the tape inside,'' it adds a spatial instruction to move higher, eliciting a motion that avoids the observed collision. These examples illustrate how rollout feedback helps refine the instruction to elicit a more effective behavior from the policy.
}

\subsection{Applications to Language Model Tasks}
\label{subsec:exp:llm}

We also show the generality of PDE beyond robotics by applying it to language-model RL tasks: competitive programming on LiveCodeBench~\citep{jain2025livecodebench} and mathematical reasoning on AIME 2026~\citep{dekoninck2026matharena}. In both settings, each problem $g$ has a canonical prompt $p_g$, PDE samples an alternative prompt $p \sim \rho(\cdot \mid g,\mathcal{H})$ during training, and evaluation uses only $p_g$. Full details and learning curves are provided in Appendix~\ref{app:sec:llm}.

{
\textbf{LiveCodeBench.}
We evaluate PDE with RLOO~\citep{ahmadian2024back} using $600$ training problems and held-out evaluation under canonical prompts. PDE improves accuracy at the early checkpoints, indicating faster initial learning. The advantage does not persist: the curves later converge, and RLOO finishes ahead (Figure~\ref{fig:app_lcb_pde}). The benefit in this experiment is therefore early learning efficiency rather than higher final accuracy.
}

{
\textbf{AIME 2026.}
We fine-tune Qwen3-4B~\citep{yang2025qwen3} with GRPO~\citep{shao2024deepseekmath} on all $30$ AIME 2026 problems and evaluate those same problems under canonical prompts. PDE modestly improves early accuracy and matches GRPO at the final checkpoint (Figure~\ref{fig:app_aime_pde}). This experiment measures learning on the training problems; LiveCodeBench evaluates generalization to held-out problems.
}

\section{Conclusion}

We introduced Prompt Driven Exploration (PDE), which samples prompts to explore different behaviors and holds each prompt fixed throughout the rollout. By updating a prompt distribution from rollout feedback, PDE improves sample efficiency on LIBERO-PRO and ManiSkill, solves tasks where action noise fails within the training budget, and extends to language-model RL. More broadly, PDE shows that for foundation-model policies, effective exploration can come from the context that conditions the policy, not only from perturbing its actions.

\textbf{Limitations and future work.}
PDE relies on a VLM to diagnose rollout failures, and inaccurate diagnoses may produce less useful prompts. Moreover, PDE can only elicit capabilities already present in the VLA's motor repertoire. Future work will explore enabling the VLA to generate and refine its own prompts, reducing or potentially eliminating reliance on an external supervisor.

\section*{Acknowledgments}
We thank the members of the Improbable AI Lab for their insightful comments and feedback on our initial draft. Richard Li was supported by the NSF Institute for Artificial Intelligence and Fundamental Interactions (Grant No. PHY-2019786) and the Felicis Scholars program. John Marangola's contributions to this work were supported by the Ministry of Trade, Industry, and Energy (MOTIE), Korea, through the Global Industrial Technology Cooperation Center program, supervised by the Korea Institute for Advancement of Technology (KIAT) (Grant No. P0028435). Nitish Dashora was supported by the National Science Foundation Graduate Research Fellowship (Grant No. 2141064).

\section*{Author Contributions}
\begin{itemize}[leftmargin=*]
    \item \textbf{Sunshine Jiang:} Led the project and manuscript writing; implemented the method; conducted both simulated and real-world experiments and built their training pipeline; conceived the mixed back-propagation method; created all figures; and conceived and implemented the illustrative example.
    \item \textbf{John Marangola:} Conducted real-world experiments, contributed to the training pipeline and architecture, and built infrastructure for data collection and inference. Co-designed the real-world experiments. Contributed to writing the manuscript, including the related work and introduction, and to revising the results and real-world sections.

    \item \textbf{David Zhang:} Conducted VLA experiments involving RND and Robometer.
    \item \textbf{Raghuram Kowdeed:} Conducted the LiveCodeBench experiments.
    \item \textbf{Ruiyang Luo:} Conducted the AIME experiments.
    \item \textbf{Nitish Dashora:} Collected real-world data and contributed to the related work section.
    \item \textbf{Richard Li:} Collected real-world data and contributed to the related work section.
    \item \textbf{Pulkit Agrawal:} Advised on the framing of the conceptual contribution.
    \item \textbf{Zhang-Wei Hong:} Conceived the idea of using prompts to drive exploration, coordinated the project, and contributed to manuscript writing.
\end{itemize}

\subsection*{AI use statement}
Generative AI tools were used to assist with proofreading, manuscript editing, LaTeX formatting, and conversion to the NeurIPS 2026 camera-ready template. The authors reviewed all AI-assisted changes and take responsibility for the final content.

\bibliographystyle{abbrv}
\bibliography{references}

\clearpage\appendix\renewcommand{\thetable}{A\arabic{table}}\renewcommand{\thefigure}{A\arabic{figure}}

\section{Implementation Details}
\label{app:impl}

\subsection{VLM Prompt Sampler}
\label{app:supervisor}

In the VLA implementation, PDE uses a VLM supervisor $\mathcal{V}$ to propose candidate prompts during discovery. During policy optimization, the exploratory sampler $\rho$ is uniform over the discovered prompt pool. As in Section~\ref{sec:framework}, the interaction
history before iteration $t$ is
\begin{equation}
\mathcal{H}_t
=
\{(g_i,p_i,\tau_i,R(\tau_i,g_i))\}_{i<t},
\end{equation}
where $g_i$ is the task, $p_i$ is the prompt used for rollout, $\tau_i$ is the
resulting trajectory, and $R(\tau_i,g_i)$ is the task reward.

In implementation, we store a compact task-specific history to fit the VLM
context window. For each task $g$, we define
\begin{equation}
\widetilde{\mathcal{H}}_t^g
=
\{(p_i,c_i,s_i)\;:\; g_i=g,\; i<t\},
\end{equation}
where $s_i \in [0,1]$ is the empirical success rate of prompt $p_i$ over $N$
rollouts, and $c_i$ is a one-sentence VLM summary of the corresponding rollout
behavior. Thus, $\widetilde{\mathcal{H}}_t^g$ is a compressed representation of
the full history $\mathcal{H}_t$ restricted to task $g$.

We also maintain a task-specific prompt pool
\begin{equation}
\mathcal{P}_t^g
\subseteq
\{p_i : (p_i,c_i,s_i)\in \widetilde{\mathcal{H}}_t^g\},
\end{equation}
containing prompts with positive empirical success rates ($s>0$). At
prompt-update iteration $t$, we query the VLM as
\begin{equation}
p \sim \mathcal{V}(\cdot \mid p_g,\mathcal{P}_t^g,\widetilde{\mathcal{H}}_t^g),
\end{equation}
where $p_g$ is the canonical prompt for task $g$. The VLM proposes candidates conditioned on the canonical prompt, the successful prompt pool, and compressed rollout feedback. Prompt refinement is implemented by appending the latest rollout
summaries and success rates to $\widetilde{\mathcal{H}}_t^g$ and re-querying
the VLM.

\paragraph{Rollout summaries.}
At each prompt-update iteration, a single call to $\mathcal{V}$ takes the rollout videos as input and jointly summarizes the behavior under each prompt and proposes new candidate prompts. Each summary $c_i$ describes in one sentence what the policy attempted and where it failed. We retain these summaries in the task-specific history rather than the videos, so the stored history grows with the number of evaluated prompts rather than the number of frames. Appendix~\ref{app:prompts} lists the combined prompt template and system prompt.

\paragraph{Pool admission.}
At each prompt-update iteration, we evaluate $K$ candidate prompts. A candidate
$p$ is added to $\mathcal{P}_t^g$ if its empirical success rate satisfies
$s>0$. Prompts with $s=0$ are still stored in
$\widetilde{\mathcal{H}}_t^g$ as negative feedback. Prompts with perfect success
are kept in the pool without further refinement.

\subsection{Prompt Discovery and Policy Optimization Schedule}
\label{app:schedule}

The VLA experiments use the discover-and-train special case of the general
framework: prompt refinement stops after the required bootstrapping stage, and the sampler remains fixed during policy optimization. Thus, the prompt sampler is refined against
a fixed initial policy. This avoids a
moving target: if $\theta$ changes while prompts are being evaluated, the same
prompt may appear good or bad for reasons unrelated to the prompt update.

We use a two-stage schedule:
\begin{itemize}[leftmargin=*, itemsep=2pt, topsep=2pt]
    \item \textbf{Prompt discovery.} We freeze $\theta$. For each task
    $g$, PDE queries $\mathcal{V}$ for candidate prompts, evaluates each prompt for $N$
    rollouts, and updates $\widetilde{\mathcal{H}}_t^g$ and
    $\mathcal{P}_t^g$.

    \item \textbf{RL update.} We freeze the discovered prompt pools and optimize
    $\theta$ with PPO. Rollouts are collected from a mixture of the canonical
    prompt $p_g$ and the discovered prompt pool $\mathcal{P}^g$, as described
    below.
\end{itemize}
We run bootstrapping for 10 prompt-update iterations, then freeze the discovered prompt pools and begin policy optimization.

\subsection{Anchoring Policy Updates to the Canonical Prompt}
\label{app:anchoring}

During RL, exploratory prompts provide successful rollouts, but evaluation is
always under the canonical prompt $p_g$. We therefore use two mechanisms to
transfer improvements from exploratory prompts back to $p_g$.

\paragraph{Mixture sampling.}
For task $g$, each rollout prompt is sampled as
\begin{equation}
p \sim \alpha_t \delta_{p_g}
+
(1-\alpha_t)\mathrm{Uniform}(\mathcal{P}^g),
\end{equation}
where $\mathcal{P}^g$ is the frozen prompt pool found during prompt discovery.
Prompts from $\mathcal{P}^g$ provide reward signal early in training, while
rollouts under $p_g$ keep optimization aligned with evaluation.

\paragraph{Adaptive mixture coefficient.}
Let $s_t^{(p_g)}$ be the empirical success rate of rollouts collected under the
canonical prompt at training step $t$. We initialize the exponential moving average at $\bar{s}_0=0.5$. The mixture coefficient $\alpha_t$ uses the EMA from the previous batch; after collecting batch $t$, we update the EMA:
\begin{align}
\bar{s}_t &= \beta s_t^{(p_g)} + (1-\beta)\bar{s}_{t-1}, \\
\alpha_t &= \mathrm{clip}\!\left(\frac{\bar{s}_{t-1}}{c},\alpha_{\min},1\right),
\end{align}
where $\beta$ is the smoothing factor, $c$ is the consolidation target, and
$\alpha_{\min}>0$ ensures that the canonical prompt is always sampled. When
canonical-prompt success is low, most rollouts come from the discovered prompt
pool. As canonical-prompt success improves, $\alpha_t$ increases and training
shifts toward $p_g$.

\paragraph{Mixed backpropagation.}
A rollout collected under an exploratory prompt $p$ directly updates
$\pi_\theta(\cdot \mid o,p)$, but evaluation uses
$\pi_\theta(\cdot \mid o,p_g)$. To couple these two prompt-conditioned policies,
we replace the current-policy log-probability in the PPO ratio with
\begin{equation}
\ell_t(\theta)
=
\frac{1}{2}\log \pi_\theta(a_t \mid o_t,p_g)
+
\frac{1}{2}\log \pi_\theta(a_t \mid o_t,p).
\end{equation}
The PPO ratio becomes
\begin{equation}
w_t(\theta)
=
\exp\!\left(
\ell_t(\theta)
-
\log \pi_{\mathrm{old}}(a_t \mid o_t,p)
\right),
\end{equation}
where the denominator uses the old log-probability recorded under the rollout
prompt $p$. When clipping is inactive, the same advantage contributes gradients under both prompts, reinforcing actions with positive advantage and suppressing those with negative advantage. This directly trains the canonical-prompt policy on exploratory rollouts.

The VLA experiments use this mixed update for $\pi_0$ and $\pi_{0.5}$. GR00T instead computes policy gradients only under the prompt used to collect the rollout; its prompt-sampling schedule is unchanged.

\subsection{Hyperparameters}
\label{app:hparams}

Table~\ref{tab:hparams} lists hyperparameters for the LIBERO and LIBERO-PRO VLA experiments. LLM settings are provided in Appendix~\ref{app:sec:llm}.

\begin{table}[t]
\centering
\small
\caption{Hyperparameters for PDE. Values in the top two blocks are specific to PDE; the bottom block lists the standard PPO hyperparameters used for policy-gradient updates during the RL stage.}
\label{tab:hparams}
\renewcommand{\arraystretch}{1.15}
\setlength{\tabcolsep}{8pt}
\begin{tabular}{@{}llc@{}}
\toprule
\textbf{Symbol} & \textbf{Description} & \textbf{Value} \\
\midrule
\multicolumn{3}{@{}l}{\textit{Prompt-discovery loop}} \\
\midrule
$T_0$        & Pool-building iterations                      & 10 \\
$K$          & Candidate prompts per iteration               & 5 \\
$N$          & Rollouts per candidate                        & 10 \\
---          & Pool admission criterion                      & $s>0$ \\
\midrule
\multicolumn{3}{@{}l}{\textit{Deployment-prompt anchoring}} \\
\midrule
$\alpha_{\min}$ & Floor on deployment-prompt mixture weight  & 0.05 \\
$\beta$         & EMA smoothing factor                       & 0.3 \\
$c$             & Consolidation target                       & 0.5 \\
\midrule
\multicolumn{3}{@{}l}{\textit{PPO}} \\
\midrule
$\epsilon$      & Clip range                                 & 0.2 \\
$\eta_{\text{lr}}$ & Learning rate                           & $5 \times 10^{-6}$ \\
$B$             & Batch size                                 & 2048 \\
$\lambda$       & GAE $\lambda$                              & 0.95 \\
$\gamma$        & Discount factor                            & 0.99 \\
$E$             & Epochs per update                          & 4 \\
\bottomrule
\end{tabular}
\end{table}

\subsection{Supervisor Prompt Templates}
\label{app:prompts}

Below are two supervisor templates: (i)~a combined candidate-generation and rollout-summarization prompt, and (ii)~the system prompt. Placeholders appear in \texttt{\{curly braces\}}.

\paragraph{(i) Candidate generation and rollout summarization}
\begin{quote}\small\ttfamily
\#\# Training Step \{step\} --- Prompt Pool

**GOAL: \{task\_description\}**\\
**Original prompt**: "\{task\_description\}" | EMA success: \{orig\_rate\} [window trend: \{trend\}]

\#\#\# Current Pool Prompts and Their Rollout Videos

**Prompt**: "\{prompt\_i\}" | EMA success: \{ema\_i\}\\
\{rollout video or frames for prompt\_i\}

\dots\ (repeated for each prompt in the pool)

\#\#\# Your Task

1. **Analyze each video**: What is the robot doing under each prompt? Which prompts produce the best behavior toward the goal?

2. **Summarize each prompt's effect** in one line.

3. **Generate 5 NEW candidate prompts** for evaluation:\\
\quad - Address failure modes you observed\\
\quad - Be semantically diverse from existing prompts\\
\quad - Try different phrasings, action verbs, spatial references

Return ONLY valid JSON:\\
\{\\
\quad "new\_prompts": ["new prompt 1", "new prompt 2", "new prompt 3", "new prompt 4", "new prompt 5"],\\
\quad "summaries": \{"exact prompt text": "one-line behavior summary", \dots\},\\
\quad "analysis": "brief reasoning about what to try next"\\
\}
\end{quote}

\paragraph{(ii) System prompt}
\begin{quote}\small\ttfamily
You are an expert in robotic manipulation and prompt engineering for vision-language-action models.

Each prompt is accompanied by \{frames\_per\_video\} frames from a robot rollout video, shown in temporal order (Frame 1 = start, Frame \{frames\_per\_video\} = end). Trace the robot's movement across frames to understand what it does over time.

Your tasks:\\
1. Analyze the rollout videos --- what is the robot doing? How close to the goal?\\
2. Identify failure modes (wrong object, wrong location, failed grasp, etc.)\\
3. Analyze how different prompts affect the robot's behavior\\
4. Generate 5 NEW candidate prompts for evaluation

The GOAL is: "\{task\_description\}"\\
The robot is controlled by a vision-language-action model ($\pi_{0.5}$).

Guidelines for new prompts:\\
- Be clear and specific about the object and target location\\
- Use natural language similar to LIBERO benchmark conventions\\
- Try different levels of specificity, action verbs, spatial references\\
- Keep prompts concise (5--15 words typically)
\end{quote}

\section{Theoretical Analysis of Mixed Back-Propagation}
\label{app:mixed-backprop}

Mixed back-propagation applies each rollout advantage to both the exploratory and canonical prompt conditions through a geometric-mean likelihood ratio.

\paragraph{Objective.}
Let \(p\) be the exploratory prompt used to collect \(a_t\), \(p_g\) the canonical prompt, and \(Q_t(a)=\pi_{\theta_{\mathrm{old}}}(a\mid o_t,p)\). Define
\[
r_t^{p}(\theta)=\frac{\pi_\theta(a_t\mid o_t,p)}{Q_t(a_t)},\qquad
r_t^{g}(\theta)=\frac{\pi_\theta(a_t\mid o_t,p_g)}{Q_t(a_t)},
\]
and
\[
r_t^{\mathrm{mix}}(\theta)=\sqrt{r_t^{p}(\theta)r_t^{g}(\theta)}
=\frac{\sqrt{\pi_\theta(a_t\mid o_t,p)\pi_\theta(a_t\mid o_t,p_g)}}{Q_t(a_t)}.
\]
The clipped surrogate is
\[
\mathcal{L}_{\mathrm{mix}}(\theta)=
\mathbb{E}_{a_t\sim Q_t}\!\left[
\min\!\left(
r_t^{\mathrm{mix}}(\theta)\hat A_t,
\operatorname{clip}\!\left(r_t^{\mathrm{mix}}(\theta),1-\epsilon,1+\epsilon\right)\hat A_t
\right)\right].
\]
When clipping is inactive,
\[
\nabla_\theta\log r_t^{\mathrm{mix}}(\theta)
=\frac{1}{2}\nabla_\theta\log\pi_\theta(a_t\mid o_t,p)
+\frac{1}{2}\nabla_\theta\log\pi_\theta(a_t\mid o_t,p_g).
\]
Thus, the same advantage updates the sampled action under both prompt conditions. Because the geometric mean is not generally normalized, this is a coupled surrogate rather than an unbiased importance-sampling estimator of either single-prompt policy gradient.

\paragraph{Local likelihood-weight analysis.}
Let \(P_t(a)=\pi_{\theta_{\mathrm{old}}}(a\mid o_t,p_g)\) and \(d_t(a)=P_t(a)/Q_t(a)\), assuming \(P_t\) is absolutely continuous with respect to \(Q_t\). At \(\theta=\theta_{\mathrm{old}}\),
\[
r_t^{p}=1,\qquad r_t^{g}=d_t(a_t),\qquad r_t^{\mathrm{mix}}=\sqrt{d_t(a_t)}.
\]
Unlike the standard PPO ratio, $r_t^{\mathrm{mix}}$ need not equal one before any parameter update. The clipped branch can therefore already be active when $\hat A_t>0$ and $r_t^{\mathrm{mix}}>1+\epsilon$, or when $\hat A_t<0$ and $r_t^{\mathrm{mix}}<1-\epsilon$. Clipping thus acts on a surrogate that reflects both prompt mismatch and policy change.
Direct canonical weighting satisfies
\[
\mathbb{E}_{a\sim Q_t}[d_t(a)^2]
=\exp\!\left(D_2(P_t\Vert Q_t)\right),\qquad
\operatorname{Var}_{Q_t}[d_t]
=\exp\!\left(D_2(P_t\Vert Q_t)\right)-1,
\]
where
\[
D_2(P_t\Vert Q_t)
=\log\int\frac{P_t(a)^2}{Q_t(a)}\,da
\]
is the order-2 Rényi divergence, which characterizes the second moment of the importance weight \(P_t/Q_t\).

In contrast,
\[
\mathbb{E}_{a\sim Q_t}\!\left[\left(\sqrt{d_t(a)}\right)^2\right]=1,\qquad
\mathbb{E}_{a\sim Q_t}[\sqrt{d_t(a)}]
=\operatorname{BC}(P_t,Q_t),
\]
where
\[
\operatorname{BC}(P_t,Q_t)
=\int\sqrt{P_t(a)Q_t(a)}\,da
\]
is the Bhattacharyya coefficient, which measures distributional overlap and lies in \([0,1]\). Therefore,
\[
\operatorname{Var}_{Q_t}[\sqrt{d_t}]
=1-\operatorname{BC}(P_t,Q_t)^2\leq 1.
\]
The square-root ratio therefore avoids the potentially exponential second moment of direct canonical weighting at the beginning of each PPO update. This bound applies only to the likelihood multiplier at \(\theta_{\mathrm{old}}\), not to the complete stochastic-gradient variance.

\paragraph{Empirical diagnostics.}
We measure
\[
\frac{\pi_\theta(a_t\mid o_t,p_g)}{\pi_\theta(a_t\mid o_t,p)}
\qquad\text{and}\qquad
D_{\mathrm{KL}}\!\left(
\pi_\theta(\cdot\mid o_t,p)\Vert
\pi_\theta(\cdot\mid o_t,p_g)
\right).
\]
We compute both diagnostics over the observation--action pairs and prompts in each rollout batch, averaging within the batch. Table~\ref{tab:mixed-backprop-diagnostics} reports these batch averages for $\pi_{0.5}$ across 12 LIBERO-PRO suites. We estimate the KL using the k1 estimator, the sample mean of $\log\pi_\theta(a_t\mid o_t,p)-\log\pi_\theta(a_t\mid o_t,p_g)$ for actions sampled under $p$.
\begin{table}[t]
\centering
\small
\caption{Likelihood-ratio and prompt-conditioned KL diagnostics, reported as mean \(\pm\) standard error across 12 LIBERO-PRO suites.}
\label{tab:mixed-backprop-diagnostics}
\begin{tabular}{c cc}
\toprule
Training step & Mean likelihood ratio & Prompt-conditioned KL \\
\midrule
0   & \(0.9128 \pm 0.0113\) & \(1.212 \pm 0.612\) \\
20  & \(0.9671 \pm 0.0083\) & \(1.191 \pm 0.694\) \\
40  & \(0.9701 \pm 0.0091\) & \(0.624 \pm 0.314\) \\
60  & \(0.9792 \pm 0.0067\) & \(0.430 \pm 0.180\) \\
80  & \(0.9712 \pm 0.0076\) & \(0.255 \pm 0.123\) \\
100 & \(0.9805 \pm 0.0081\) & \(0.334 \pm 0.203\) \\
120 & \(0.9815 \pm 0.0093\) & \(0.336 \pm 0.159\) \\
\bottomrule
\end{tabular}
\end{table}
The mean likelihood ratio increases from \(0.9128\) to \(0.9671\) by step \(20\) and remains between \(0.9671\) and \(0.9815\) thereafter. The KL decreases overall from \(1.212\) to \(0.336\), showing that the mismatch between the two prompt-conditioned distributions decreases during training.
Mixed back-propagation achieves a training success rate of \(59.6\%\), compared with \(49.6\%\) for canonical-only back-propagation and \(45.0\%\) for exploratory-only back-propagation.

\section{Experiment Details}
\label{app:exp}

\subsection{Task Difficulty Groups}
\label{app:difficulty-groups}
For simulation policy evaluation, we use $250$ evaluation environments per task. LIBERO-PRO covers $120$ task instances each for $\pi_{0.5}$ and GR00T and $100$ for $\pi_0$; the $\pi_0$ evaluation omits task perturbations for the Object and LIBERO-10 suites. We exclude language perturbations because PDE itself varies prompts, and environment perturbations because their BDDL files are unavailable.

We group tasks by initial success. For $\pi_{0.5}$, hard, medium, and easy tasks have success below $10\%$, in $[10\%,80\%)$, and at least $80\%$, respectively, with $47$, $35$, and $38$ tasks. For GR00T and $\pi_0$, the corresponding thresholds are zero, $(0,40\%]$, and above $40\%$; task counts are $49/31/40$ for GR00T and $49/24/27$ for $\pi_0$.

For ManiSkill, hard tasks have initial success below $25\%$, medium tasks have initial success in $[25\%,50\%)$, and easy tasks have initial success of at least $50\%$.

\subsection{ManiSkill Setup}
\label{app:maniskill-setup}
{
Following Liu et al.~\citep{liu2026what}, we use ManiSkill \texttt{PutOnPlateInScene25}~\citep{tao2024maniskill3}, where the policy places one of $25$ objects onto a receptacle in a kitchen scene. The $\pi_{0.5}$ SFT checkpoint is trained only on the in-distribution split, which contains $16$ objects and $16{,}384$ episodes synthesized with MPLib\footnote{\url{https://github.com/haosulab/mplib}}. We fine-tune one policy per OOD variant with PPO and evaluate on that variant. These shifts are relative to the SFT distribution, so the protocol measures adaptation within each shifted setting. The $12$ variants comprise five visual shifts (backgrounds and textures modified by image overlays of varying intensity), four semantic shifts (unseen objects, varied instructions, and distractors), and three execution shifts (initial states, unseen robot poses, and dynamic disturbances).
}

\subsection{Compute}
\label{app:exp:compute}
For each experiment, we used one node (8) of H200s.  The reported 60K environment steps used to bootstrap the prompt pool
(shown in Figures~\ref{fig:agg_task_difficulty},
\ref{fig:agg_task_difficulty_pi0}, and
\ref{fig:agg_task_difficulty_gr00t})
and 10 VLM calls are totals for an entire suite.
Each suite contains 10 tasks, each specified by a canonical instruction.
Prompt-pool construction proceeds for 10 suite-level iterations. At each
iteration, a single VLM call jointly diagnoses failures and proposes
updated prompts for all 10 tasks. The reported cost therefore includes
all prompt-update iterations and associated rollouts across the full
suite.

Once the prompt pool has been constructed, it remains fixed throughout
PPO training. PDE only updates the sampling distribution over this fixed
pool using standard PPO rollouts. It therefore requires no additional
VLM calls or environment interactions beyond the 60K pool-construction
steps and the standard PPO budget.

Table~\ref{tab:compute_cost} summarizes the environment-interaction and
compute costs of PDE and the main baselines. Robometer requires two
8-H200 nodes---one for policy training and one for hosting the Robometer
model---whereas PDE and the remaining baselines use a single 8-H200
node.

\begin{table}[t]
    \centering
    \caption{Environment-interaction and compute costs for PDE and the
    main baselines. Prompt-pool construction costs are reported for an
    entire 10-task suite.}
    \label{tab:compute_cost}
    \resizebox{\linewidth}{!}{
    \begin{tabular}{lcccccc}
        \toprule
        \textbf{Method}
        & \textbf{PPO Steps}
        & \textbf{Pool Construction Steps}
        & \textbf{Total Env. Steps}
        & \textbf{VLM Calls}
        & \textbf{Wall-Clock Time}
        & \textbf{H200 GPUs} \\
        \midrule
        PDE
        & 15.00M
        & 0.06M
        & 15.06M
        & 10
        & 72.5 h
        & 8 \\
        PPO / Action Noise / Parameter Noise / RND
        & 15.00M
        & 0
        & 15.00M
        & 0
        & 72 h
        & 8 \\
        Robometer
        & 15.00M
        & 0
        & 15.00M
        & 0
        & 80 h
        & 16 \\
        \bottomrule
    \end{tabular}
    }
\end{table}

Relative to the 15M-step PPO budget, constructing the prompt pool adds
only 60K environment steps, corresponding to a \(0.4\%\) increase in
environment interactions, and approximately \(0.5\) hours of wall-clock
time. Despite this small overhead, PDE reaches the final performance of
Action Noise using 6--9M fewer PPO steps, corresponding to approximately
\(1.7\)--\(2.5\times\) better sample efficiency. The cost of constructing
the prompt pool is therefore too small to explain the observed
performance improvement.

\subsection{Real-World Experiment Details}
\label{sec:exp-real}

\begin{figure}[!htbp]
    \centering
    \includegraphics[width=0.55\textwidth]{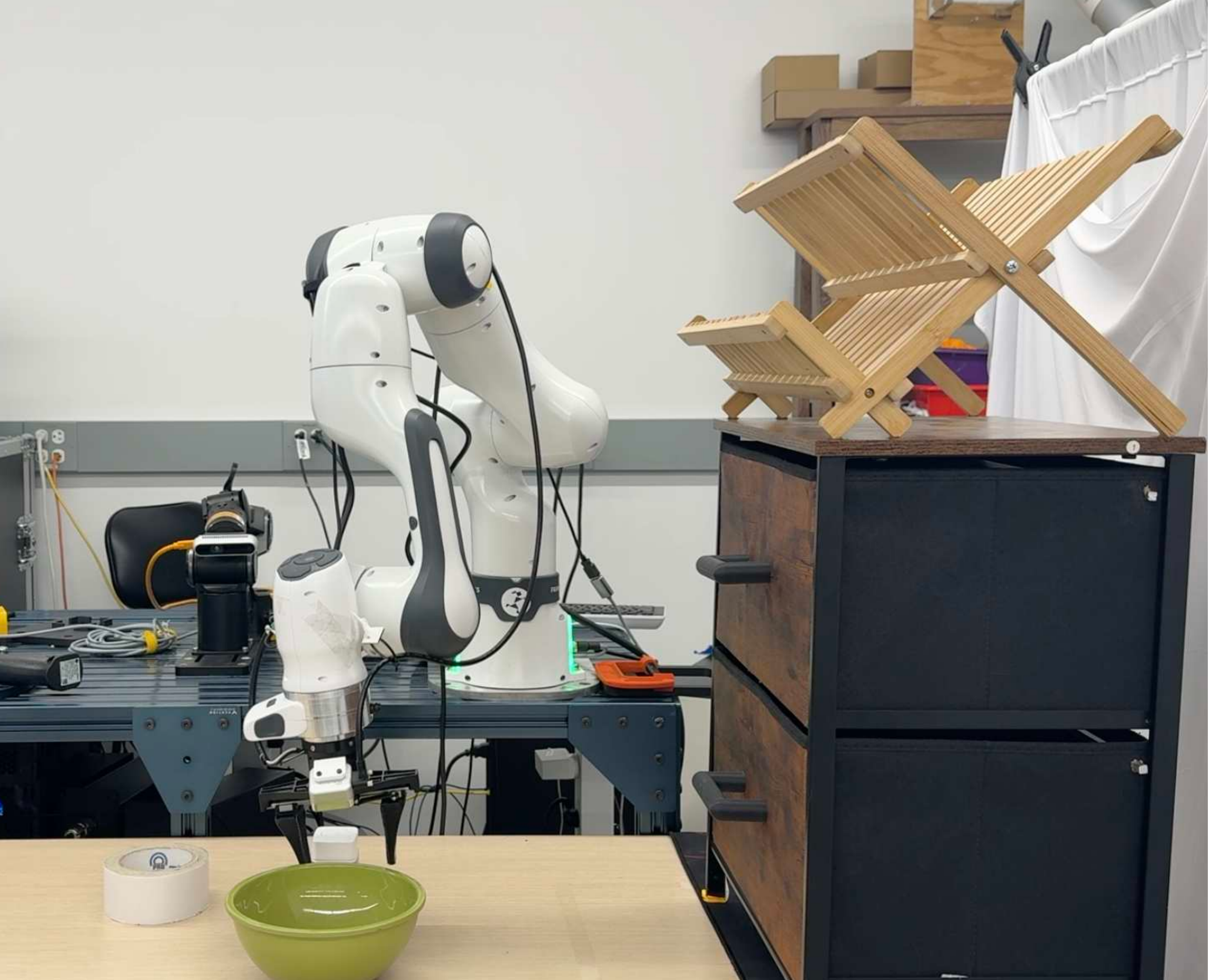}
    \caption{Real-world experiment setup.}
    \label{fig:setup-realworld}
\end{figure}

\subsubsection{SFT Task Suite}
\label{app:sft_tasks}

The $\pi_{0.5}$ checkpoint used as the starting point for all RL fine-tuning experiments is trained on the ten language-conditioned manipulation tasks listed in Table~\ref{tab:sft_tasks}. The tasks share a fixed tabletop scene containing four manipulable objects (tape, bowl, green container, and a small object placed on a marked region of the workspace), two stacked drawers (top and bottom), and a two-level rack. Together they span four object--receptacle relationships---placing an object \emph{in} a bowl (task 1), \emph{on} a rack (tasks 2, 3, 10), \emph{in front of} the camera (tasks 4, 5), and \emph{inside} a drawer (tasks 6--9)---over three target objects (tape, bowl, green container) and six receptacle locations (bowl, top rack, bottom rack, top drawer, bottom drawer, front of the camera).

Each task is collected via teleoperated demonstrations on the Franka FR3 platform described in Section~\ref{subsec:exp:real}, with 40 demonstrations per task and object positions randomized within a fixed workspace region across demonstrations. We hold the demonstration budget fixed across tasks rather than scaling to per-task difficulty, so the SFT checkpoint's per-task competence reflects the intrinsic difficulty of each task under uniform supervision.

\begin{table}[t]
    \centering
    \caption{\textbf{SFT task suite.} The ten tasks the $\pi_{0.5}$ SFT checkpoint is trained on, with 40 teleoperated demonstrations per task. }
    \label{tab:sft_tasks}
    \small
    \begin{tabular}{cl}
        \toprule
        \textbf{\#} & \textbf{Prompt} \\
        \midrule
        1  & ``\textit{put the green container in the bowl}'' \\
        2  & ``\textit{put the tape on the top rack}'' \\
        3  & ``\textit{put the bowl on the top rack}'' \\
        4  & ``\textit{put the green container in front of the camera}'' \\
        5  & ``\textit{put the bowl in front of the camera}'' \\
        6  & ``\textit{open the top drawer and put the bowl inside}'' \\
        7  & ``\textit{open the bottom drawer and put the bowl inside}'' \\
        8  & ``\textit{open the top drawer and put the tape inside}'' \\
        9  & ``\textit{open the top drawer and put the green container inside}'' \\
        10 & ``\textit{put the green container on the bottom rack}'' \\
        \bottomrule
    \end{tabular}
\end{table}

\subsubsection{Detailed Results}
Table~\ref{tab:sr} reports per-task success rates on the three real-world
manipulation tasks. All methods start from the same SFT checkpoint and are
given an identical online-interaction budget of 128 real-world rollouts per task (two
iterations of 64 rollouts each); every entry is measured over 20 evaluation
rollouts. Under this small budget, vanilla PPO yields only modest gains and degrades
one task, while PPO+PDE improves every task by +20.0 to +40.0 points,
tripling the average success rate of the SFT policy (15.0\% to 45.0\%) and
more than doubling that of vanilla PPO.

\begin{table*}[t]
\centering
\caption{\textbf{Per-task success rate (\%) on real-world tasks.}
\textit{SFT}: frozen checkpoint before any RL fine-tuning; \textit{PPO},
\textit{PPO+PDE}: final success rates after RL fine-tuning with a budget of
128 real-world rollouts per task (2 iterations of 64), over 20 eval rollouts per task;
(+X) is the gain in percentage points over SFT. Vanilla PPO yields only
modest gains and even degrades one task; adding PDE improves every task and
attains the highest final success rate across the board, tripling the average
success rate of the SFT policy.}
\label{tab:sr}
\resizebox{0.9\textwidth}{!}{%
\begin{tabular}{l|c|c|>{\columncolor{ourskyblue}}c}
\toprule
\textbf{Task} & \textbf{SFT} & \textbf{PPO} & \multicolumn{1}{c}{\begin{tabular}{c}\textbf{PPO + PDE}\\\textbf{(Ours)}\end{tabular}} \\
\midrule
``place the green container on the bottom rack'' & 20.0 & 15.0 ($-$5.0) & \textbf{60.0} (+40.0) \\
``open the top drawer and put the tape inside''  & 15.0 & 30.0 (+15.0)  & \textbf{45.0} (+30.0) \\
``put the green container in the bowl''          & 10.0 & 15.0 (+5.0)   & \textbf{30.0} (+20.0) \\
\midrule
\rowcolor{gray!20}
\textbf{Average}                                 & 15.0 & 20.0 (+5.0)   & \textbf{45.0} (+30.0) \\
\bottomrule
\end{tabular}%
}
\end{table*}

\subsection{Ablation Studies}
\label{sec:exp-ablation}

\textbf{Tasks in the illustrative learning comparison.} Regimes 1--3 in Table~\ref{tab:libero-regime-breakdown} use task IDs 0, 5, and 2, respectively, from LIBERO-PRO Object with task perturbations. Each regime illustrates the learning behavior on one selected task.

\noindent\textbf{Is prompt optimization necessary, or do random paraphrases suffice?}
A natural concern is whether PDE's gains come simply from exposing the policy to \emph{diverse}
phrasings of the goal, rather than from \emph{optimizing} which phrasings enter the curriculum.
To isolate this, we replace PDE's discovered prompt pool with an un-optimized one: for each task we
query a VLM (Qwen3-VL-235B) for five random paraphrases of the canonical instruction $p_g$ that
describe the same task, using \emph{no} rollout feedback or success-rate filtering. Everything
else---the EMA mixed-backpropagation curriculum, the consolidation floor, and all
hyperparameters---is held fixed. We compare this \emph{Random Prompts} variant against PDE
(optimized pool) and a plain PPO baseline trained only on $p_g$, on the Object suite with task
perturbation.

\begin{figure}[t]
\centering
\includegraphics[width=0.8\columnwidth]{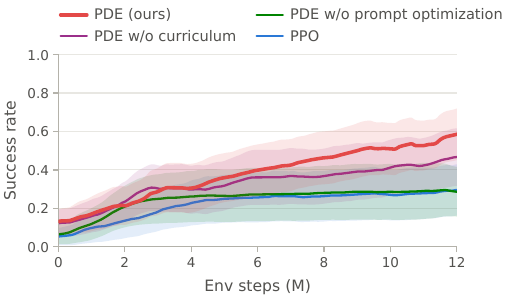}
\caption{\textbf{Prompt discovery and adaptive sampling both contribute.} Replacing discovered prompts with random paraphrases or removing the adaptive sampling schedule reduces success on LIBERO-PRO Object with task perturbations. Curves show mean training success across tasks; shading denotes $\pm 1$ standard error.}
\label{fig:random-prompt-baseline}
\end{figure}

Figure~\ref{fig:random-prompt-baseline} shows that random paraphrases improve training success over PPO but remain below PDE. Rollout-guided prompt refinement therefore provides benefits beyond exposing the policy to varied phrasings. The
reason is visible at the level of individual prompts: evaluated on the base SFT policy, the random
paraphrases attain a mean success rate of just $6.8\%$ on the tasks where PDE's optimized prompts
reach $29.2\%$, and $0\%$ on $9$ of the $10$ tasks. Un-optimized rephrasings rarely unlock reward on
the hard perturbed tasks; PDE's search is what surfaces the spatially and visually grounded phrasings
that move the policy off zero. Optimizing \emph{which} prompts enter the curriculum---not merely
diversifying them---is therefore essential.

\noindent\textbf{Does adaptive prompt sampling help?} Both full PDE and this ablation use a prompt pool constructed during bootstrapping and held fixed throughout PPO training. The ablation additionally fixes the prompt-sampling distribution, whereas full PDE adapts the mixture weight between the canonical prompt and the exploratory pool as canonical-prompt success improves. Sampling within the exploratory pool remains uniform, and the policy-update rule is unchanged.

Figure~\ref{fig:random-prompt-baseline} shows that fixed sampling improves over vanilla PPO but remains below adaptive sampling. The benefit therefore comes from adjusting prompt sampling during learning, beyond discovering useful prompts during bootstrapping.

\begin{figure}[t]
\centering
\begin{minipage}{0.48\columnwidth}
    \centering
    \includegraphics[width=\linewidth]{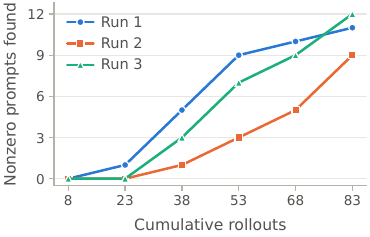}
    \captionof{figure}{Cumulative unique nonzero prompts discovered vs.\ cumulative rollouts across three independent runs on the microwave task.}
    \label{fig:prompt_discovery}
\end{minipage}
\hfill
\begin{minipage}{0.48\columnwidth}
    \centering
    \small
    \setlength{\tabcolsep}{4pt}
    \renewcommand{\arraystretch}{0.95}
    \captionof{table}{\small Success Rate (\%) on the Microwave task across $10$ prompts ($5$ IND, $5$ OOD).}
    \label{tab:microwave-prompt-sensitivity}
    \begin{tabular}{lcc}
    \toprule
    & \textbf{IND Prompts}  & \textbf{OOD Prompts}  \\
    \midrule
    Action Noise & $11.2{\scriptstyle \pm 2.9}$ & $10.4{\scriptstyle \pm 3.0}$ \\
    \rowcolor{ourskyblue}
    PDE (Ours) & $\mathbf{88.8}{\scriptstyle \pm 2.0}$ & $\mathbf{95.2}{\scriptstyle \pm 2.9}$ \\
    \bottomrule
    \end{tabular}
\end{minipage}
\end{figure}

\noindent\textbf{How does the number of discovered prompts scale with the search budget?} Prompt discovery is both consistent across runs and scales reliably with the search budget: the cumulative number of nonzero prompts grows approximately linearly with rollouts, with each independent run finding $9$--$12$ nonzero prompts after $84$ rollouts (Figure~\ref{fig:prompt_discovery}). This suggests that allocating more rollouts directly translates to a larger and more diverse curriculum pool.

\noindent\textbf{Does PDE produce a paraphrase-robust policy?} PDE trains under a learned distribution over prompts, so the resulting policy is exposed to many surface realizations of the same goal during fine-tuning. We construct two prompt sets of size $5$: \emph{(i) IN-Distribution Train Prompts}---the canonical instruction $p_g{=}$\emph{``close the microwave''} together with the four exploratory phrasings PDE drew from $\rho(\cdot \mid g, \mathcal{H})$ during training; \emph{(ii) Out-Of-Distribution Paraphrases}---five LLM-generated rephrasings that the policy never saw. As shown in Table~\ref{tab:microwave-prompt-sensitivity}, PDE generalizes across the surface form of the instruction, with no drop on unseen prompts.

\begin{wraptable}{r}{0.3\textwidth}
\vspace{-10pt}
\centering\small
\caption{Eval SR on original prompt.}
\label{tab:curriculum_deploy}
\begin{tabular}{lc}
\toprule
Method & SR \\
\midrule
Best Prompt (no curr.) & 0.396 \\
Prompt Curriculum & \textbf{0.638} \\
\bottomrule
\end{tabular}
\vspace{-10pt}
\end{wraptable}

\noindent\textbf{Is a single best prompt enough?}
PDE mixes rollouts from the discovered prompt pool and gradually consolidates training toward the canonical prompt $p_g$. A natural alternative is to train exclusively on the highest-success discovered prompt or to deploy that prompt instead of $p_g$.

These settings correspond to different deployment assumptions. In terms of training success, the single-best-prompt variant achieves $65.4\%$ success, compared with $59.6\%$ for the default full-pool configuration (Table~\ref{tab:ablation-object-task}), because it can specialize entirely to an optimized non-canonical prompt. These results are therefore not directly comparable under canonical-prompt deployment.

When both methods are instead evaluated under the original instruction $p_g$, the single-best-prompt variant drops to $39.6\%$, whereas the full-pool variant reaches $63.8\%$ (Table~\ref{tab:curriculum_deploy}), a gap of more than $24$ percentage points. The discovered prompts thus serve as scaffolds that unlock successful behavior early in training, while maintaining the full pool and consolidating toward $p_g$ transfers this behavior to the instruction used at deployment. We therefore use the full prompt pool as the default when deployment requires the canonical prompt.

\noindent\textbf{Does mixed backpropagation help?} Canonical-only updates compute gradients under $p_g$, while rollout-prompt-only updates compute gradients under the prompt used to collect each rollout. The latter includes canonical-prompt updates when a rollout is collected under $p_g$, but does not directly update the canonical condition on exploratory-prompt rollouts. Mixed backpropagation couples both conditions on those rollouts and achieves the highest training success rate (Table~\ref{tab:ablation-object-task}).

\noindent\textbf{How sensitive is the curriculum to the consolidation target $c$?} Table~\ref{tab:ablation-object-task} shows that all three settings ($c \in \{0.25, 0.5, 0.75\}$) converge to comparable final success rates within per-task SE, indicating that the method is not sensitive to this hyperparameter. We use $c{=}0.5$ as the default.  Although $c=0.75$ is numerically higher than $c=0.5$ in Table~\ref{tab:ablation-object-task} ($63.7\%$ vs.\ $59.6\%$), the differences among $c \in \{0.25, 0.5, 0.75\}$ are within the per-task standard errors, and PDE outperforms all baselines for every tested value. This indicates that the method is not particularly sensitive to the consolidation target. We use $c=0.5$ as a fixed, untuned default rather than selecting the best value post hoc, while reporting all tested settings for completeness.

\begin{table}[t]
  \centering\small
  \setlength{\tabcolsep}{6pt}
  \renewcommand{\arraystretch}{1.05}
  \caption{Training success rates for ablations on Object/task perturbation at step~120 (250 envs, mean across 10 tasks). Default variant shaded.}
  \label{tab:ablation-object-task}
  \begin{tabular}{llc}
  \toprule
  \textbf{Ablation} & \textbf{Variant} & \textbf{SR (\%)} \\
  \midrule
  Training prompts & Single best prompt & $\mathbf{65.4}$ \\
                   & Converge to Best & $63.7$ \\
                   & \cellcolor{ourskyblue}Full pool (Ours) & \cellcolor{ourskyblue}$59.6$ \\
  \midrule
  Backpropagation & Original-only & $49.6$ \\
                  & Curriculum-only & $45.0$ \\
                  & \cellcolor{ourskyblue}Mixed (Ours) & \cellcolor{ourskyblue}$\mathbf{59.6}$ \\
  \midrule
  Consolidation $c$ & $c{=}0.25$ & $58.8$ \\
                    & \cellcolor{ourskyblue}$c{=}0.5$ (Ours) & \cellcolor{ourskyblue}$59.6$ \\
                    & $c{=}0.75$ & $\mathbf{63.7}$ \\
  \bottomrule
  \end{tabular}
\end{table}

\noindent \textbf{Prompt-Only Performance Ceiling.}
To isolate the benefit of prompt optimization without RL post-training, we report the best performance achieved by PDE before any policy update. This corresponds to the first point of each training curve. As shown in Table~\ref{tab:prompt-only-ceiling}, optimized prompts substantially improve the frozen $\pi_{0.5}$ policy on hard and medium LIBERO-PRO tasks, from $1.3\%$ to $6.4\%$ and from $45.1\%$ to $54.9\%$, respectively. On easy tasks, where the canonical prompt already achieves $92.7\%$, prompt optimization provides little additional gain. These results represent the best prompt-only performance found by PDE, rather than a global upper bound over all possible prompts.

\begin{table}[t]
\centering
\caption{Prompt-only performance of the frozen $\pi_{0.5}$ policy on LIBERO-PRO, before any RL update.}
\label{tab:prompt-only-ceiling}
\begin{tabular}{lcc}
\toprule
Task difficulty & Canonical prompt & Optimized prompt \\
\midrule
Hard   & $1.3\%$  & $\mathbf{6.4\%}$ \\
Medium & $45.1\%$ & $\mathbf{54.9\%}$ \\
Easy   & $92.7\%$ & $\mathbf{93.0\%}$ \\
\bottomrule
\end{tabular}
\end{table}

\subsection{Detailed results}

\textbf{Flow Noise + Ent.} This baseline uses a learned noise network to inject stochasticity into flow-based action generation and adds an entropy bonus to encourage exploration~\citep{chen2025pirl}. Table~\ref{tab:libero-aggregated-pi05} shows that it improves hard-task success over standard Action Noise, but remains below PDE. This suggests that prompt variation provides benefits beyond learned action noise and entropy regularization.

\begin{table*}[t]
  \centering
  \footnotesize
  \setlength{\tabcolsep}{4pt}
  \renewcommand{\arraystretch}{0.95}
  \caption{\small $\pi_{0.5}$ aggregated success rate (\%) on LIBERO-PRO, by task difficulty and by perturbation axis.}
  \label{tab:libero-aggregated-pi05}
  \resizebox{\textwidth}{!}{
  \begin{tabular}{lccc|ccc|c}
  \toprule
  & \multicolumn{3}{c|}{\textbf{Difficulty}} & \multicolumn{3}{c|}{\textbf{Perturbation}} & \\
  \cmidrule{2-4} \cmidrule{5-7}
  & \textbf{Hard} & \textbf{Medium} & \textbf{Easy} & \textbf{Object} & \textbf{Swap} & \textbf{Task} & \textbf{All} \\
  \midrule
  Initial policy
  & $1.28{\scriptstyle \pm 0.40}$ & $45.08{\scriptstyle \pm 3.64}$ & $94.32{\scriptstyle \pm 0.88}$ & $61.26{\scriptstyle \pm 5.72}$ & $50.31{\scriptstyle \pm 6.55}$ & $18.99{\scriptstyle \pm 5.24}$ & $43.52{\scriptstyle \pm 3.73}$ \\
  \midrule
  Action Noise
  & $13.32{\scriptstyle \pm 4.30}$ & $78.16{\scriptstyle \pm 2.88}$ & $95.97{\scriptstyle \pm 1.06}$ & $71.97{\scriptstyle \pm 5.46}$ & $67.33{\scriptstyle \pm 6.10}$ & $35.91{\scriptstyle \pm 7.11}$ & $58.40{\scriptstyle \pm 3.87}$ \\
  Action Noise (n=0.7)
  & $17.00{\scriptstyle \pm 4.82}$ & $74.27{\scriptstyle \pm 3.65}$ & $90.53{\scriptstyle \pm 1.93}$ & $71.96{\scriptstyle \pm 5.46}$ & $64.74{\scriptstyle \pm 5.89}$ & $34.26{\scriptstyle \pm 6.56}$ & $56.99{\scriptstyle \pm 3.74}$ \\
  Parameter Noise (n=0.01)
  & $13.03{\scriptstyle \pm 4.32}$ & $77.54{\scriptstyle \pm 3.26}$ & $96.11{\scriptstyle \pm 0.64}$ & $72.89{\scriptstyle \pm 5.52}$ & $67.25{\scriptstyle \pm 6.16}$ & $34.32{\scriptstyle \pm 6.97}$ & $58.15{\scriptstyle \pm 3.90}$ \\
  Parameter Noise (n=0.1)
  & $0.77{\scriptstyle \pm 0.52}$ & $2.36{\scriptstyle \pm 1.24}$ & $18.40{\scriptstyle \pm 3.48}$ & $9.68{\scriptstyle \pm 2.83}$ & $8.39{\scriptstyle \pm 2.18}$ & $2.38{\scriptstyle \pm 1.93}$ & $6.82{\scriptstyle \pm 1.37}$ \\
  Flow Noise + Ent
  & $17.24{\scriptstyle \pm 4.80}$ & $81.08{\scriptstyle \pm 3.05}$ & $95.42{\scriptstyle \pm 0.99}$ & $75.61{\scriptstyle \pm 5.43}$ & $68.04{\scriptstyle \pm 6.07}$ & $38.20{\scriptstyle \pm 7.01}$ & $60.62{\scriptstyle \pm 3.85}$ \\
  VLAC
  & $7.07{\scriptstyle \pm 2.89}$ & $63.59{\scriptstyle \pm 4.95}$ & $93.26{\scriptstyle \pm 1.67}$ & $68.20{\scriptstyle \pm 5.73}$ & $61.75{\scriptstyle \pm 6.34}$ & $22.60{\scriptstyle \pm 5.81}$ & $50.85{\scriptstyle \pm 3.88}$ \\
  Robometer
  & $13.34{\scriptstyle \pm 4.27}$ & $77.40{\scriptstyle \pm 3.48}$ & $95.79{\scriptstyle \pm 1.50}$ & $73.33{\scriptstyle \pm 5.44}$ & $66.24{\scriptstyle \pm 6.30}$ & $34.83{\scriptstyle \pm 7.00}$ & $58.13{\scriptstyle \pm 3.91}$ \\
  RND
  & $10.53{\scriptstyle \pm 3.90}$ & $77.53{\scriptstyle \pm 2.95}$ & $90.93{\scriptstyle \pm 2.31}$ & $69.88{\scriptstyle \pm 5.37}$ & $66.17{\scriptstyle \pm 6.10}$ & $30.54{\scriptstyle \pm 6.68}$ & $55.53{\scriptstyle \pm 3.84}$ \\
  \rowcolor{ourskyblue}
  PDE (Ours)
  & $\mathbf{19.48}{\scriptstyle \pm 5.10}$ & $\mathbf{85.15}{\scriptstyle \pm 2.58}$ & $\mathbf{97.46}{\scriptstyle \pm 0.86}$ & $\mathbf{77.31}{\scriptstyle \pm 5.50}$ & $\mathbf{71.24}{\scriptstyle \pm 6.17}$ & $\mathbf{41.43}{\scriptstyle \pm 7.17}$ & $\mathbf{63.33}{\scriptstyle \pm 3.89}$ \\
  \bottomrule
  \end{tabular}
  }
\end{table*}

\begin{table*}[t]
  \centering
  \footnotesize
  \setlength{\tabcolsep}{4pt}
  \renewcommand{\arraystretch}{0.95}
  \caption{\small GR00T and $\pi_0$ aggregated success rate (\%) on LIBERO-PRO by task difficulty.}
  \label{tab:libero-aggregated-gr00t-pi0}
  \resizebox{\textwidth}{!}{
  \begin{tabular}{lcccc|cccc}
  \toprule
  & \multicolumn{4}{c|}{\textbf{GR00T}} & \multicolumn{4}{c}{\textbf{$\pi_0$}} \\
  \cmidrule{2-5} \cmidrule{6-9}
  & \textbf{Hard} & \textbf{Medium} & \textbf{Easy} & \textbf{All}
  & \textbf{Hard} & \textbf{Medium} & \textbf{Easy} & \textbf{All} \\
  \midrule

  Action Noise
  & $12.73{\scriptstyle \pm 4.24}$ & $87.23{\scriptstyle \pm 1.90}$ & $96.00{\scriptstyle \pm 0.86}$ & $59.73{\scriptstyle \pm 4.01}$
  & $12.16{\scriptstyle \pm 4.03}$ & $73.83{\scriptstyle \pm 7.05}$ & $88.44{\scriptstyle \pm 4.83}$ & $47.56{\scriptstyle \pm 4.56}$ \\
  \rowcolor{ourskyblue}
  PDE (Ours)
  & $\mathbf{20.33}{\scriptstyle \pm 5.49}$ & $\mathbf{93.68}{\scriptstyle \pm 3.16}$ & $\mathbf{98.30}{\scriptstyle \pm 0.42}$ & $\mathbf{65.27}{\scriptstyle \pm 4.17}$
  & $\mathbf{12.73}{\scriptstyle \pm 3.77}$ & $\mathbf{85.00}{\scriptstyle \pm 4.50}$ & $\mathbf{89.33}{\scriptstyle \pm 4.75}$ & $\mathbf{50.76}{\scriptstyle \pm 4.49}$ \\
  \bottomrule
  \end{tabular}
  }
\end{table*}

\begin{longtable}{@{}llllccc@{}}
\caption{Per-task success rate (\%) on LIBERO-PRO with GR00T, sorted by difficulty bucket and PPO init SR.\label{tab:libero-pertask-gr00t}}\\
\toprule
\textbf{Diff.} & \textbf{Suite} & \textbf{Pert.} & \textbf{Task} & \textbf{Init} & \textbf{PPO} & \textbf{Ours} \\
\midrule
\endfirsthead
\multicolumn{7}{c}{\textit{(Table~\ref{tab:libero-pertask-gr00t} continued)}}\\
\toprule
\textbf{Diff.} & \textbf{Suite} & \textbf{Pert.} & \textbf{Task} & \textbf{Init} & \textbf{PPO} & \textbf{EMA (Ours)} \\
\midrule
\endhead
\midrule \multicolumn{7}{r}{\textit{(continued on next page)}} \\
\endfoot
\bottomrule
\endlastfoot
easy & goal & object & 7 & 100.00 & 96.00 & \textbf{96.00} \\
easy & 10 & swap & 1 & 83.67 & 96.00 & \textbf{96.00} \\
easy & object & swap & 6 & 79.25 & 100.00 & \textbf{92.00} \\
easy & 10 & swap & 5 & 78.43 & 96.00 & \textbf{92.00} \\
easy & object & object & 8 & 78.33 & 96.00 & \textbf{100.00} \\
easy & object & swap & 8 & 78.33 & 100.00 & \textbf{100.00} \\
easy & 10 & object & 1 & 77.55 & 92.00 & \textbf{100.00} \\
easy & object & object & 4 & 76.79 & 96.00 & \textbf{100.00} \\
easy & 10 & object & 3 & 75.56 & 96.00 & \textbf{100.00} \\
easy & object & swap & 3 & 75.56 & 100.00 & \textbf{100.00} \\
easy & object & swap & 4 & 75.00 & 100.00 & \textbf{100.00} \\
easy & spatial & swap & 1 & 69.39 & 96.00 & \textbf{96.00} \\
easy & 10 & swap & 6 & 67.92 & 100.00 & \textbf{100.00} \\
easy & object & object & 6 & 67.92 & 100.00 & \textbf{100.00} \\
easy & 10 & object & 4 & 67.86 & 80.00 & \textbf{96.00} \\
easy & object & swap & 5 & 66.67 & 100.00 & \textbf{100.00} \\
easy & object & object & 1 & 63.27 & 100.00 & \textbf{100.00} \\
easy & object & swap & 1 & 63.27 & 100.00 & \textbf{100.00} \\
easy & object & object & 7 & 62.79 & 100.00 & \textbf{100.00} \\
easy & 10 & object & 2 & 62.75 & 96.00 & \textbf{100.00} \\
easy & 10 & swap & 4 & 60.71 & 88.00 & \textbf{96.00} \\
easy & object & task & 1 & 59.18 & 100.00 & \textbf{100.00} \\
easy & 10 & swap & 2 & 58.82 & 96.00 & \textbf{100.00} \\
easy & spatial & swap & 5 & 58.82 & 84.00 & \textbf{100.00} \\
easy & spatial & object & 1 & 57.14 & 84.00 & \textbf{100.00} \\
easy & object & object & 9 & 56.36 & 100.00 & \textbf{100.00} \\
easy & object & swap & 7 & 55.81 & 100.00 & \textbf{96.00} \\
easy & spatial & swap & 0 & 55.10 & 100.00 & \textbf{100.00} \\
easy & spatial & swap & 2 & 54.90 & 84.00 & \textbf{92.00} \\
easy & object & swap & 0 & 53.06 & 96.00 & \textbf{100.00} \\
easy & spatial & object & 0 & 53.06 & 100.00 & \textbf{96.00} \\
easy & spatial & object & 2 & 50.98 & 88.00 & \textbf{100.00} \\
easy & goal & swap & 5 & 49.02 & 92.00 & \textbf{100.00} \\
easy & 10 & swap & 0 & 48.98 & 92.00 & \textbf{92.00} \\
easy & object & object & 3 & 46.67 & 100.00 & \textbf{100.00} \\
easy & 10 & swap & 7 & 46.51 & 100.00 & \textbf{100.00} \\
easy & object & swap & 9 & 45.45 & 100.00 & \textbf{96.00} \\
easy & object & object & 5 & 45.10 & 100.00 & \textbf{100.00} \\
easy & goal & task & 7 & 44.19 & 100.00 & \textbf{100.00} \\
easy & goal & object & 5 & 41.18 & 96.00 & \textbf{96.00} \\
\midrule
medium & goal & swap & 8 & 38.33 & 88.00 & \textbf{88.00} \\
medium & spatial & object & 5 & 35.29 & 92.00 & \textbf{100.00} \\
medium & spatial & object & 6 & 33.96 & 96.00 & \textbf{100.00} \\
medium & spatial & object & 3 & 33.33 & 100.00 & \textbf{100.00} \\
medium & spatial & swap & 9 & 30.91 & 80.00 & \textbf{100.00} \\
medium & object & object & 2 & 29.41 & 100.00 & \textbf{100.00} \\
medium & spatial & object & 8 & 28.33 & 88.00 & \textbf{96.00} \\
medium & spatial & swap & 8 & 28.33 & 88.00 & \textbf{96.00} \\
medium & spatial & swap & 6 & 28.30 & 92.00 & \textbf{100.00} \\
medium & goal & object & 2 & 27.45 & 64.00 & \textbf{100.00} \\
medium & 10 & object & 9 & 25.45 & 64.00 & \textbf{0.00} \\
medium & spatial & object & 9 & 23.64 & 84.00 & \textbf{84.00} \\
medium & spatial & swap & 7 & 23.26 & 64.00 & \textbf{96.00} \\
medium & spatial & task & 5 & 21.57 & 92.00 & \textbf{92.00} \\
medium & spatial & task & 9 & 20.00 & 96.00 & \textbf{100.00} \\
medium & spatial & object & 4 & 19.64 & 96.00 & \textbf{96.00} \\
medium & object & swap & 2 & 19.61 & 96.00 & \textbf{100.00} \\
medium & spatial & swap & 4 & 16.07 & 96.00 & \textbf{100.00} \\
medium & 10 & object & 8 & 13.33 & 68.00 & \textbf{96.00} \\
medium & goal & object & 9 & 12.73 & 88.00 & \textbf{92.00} \\
medium & spatial & object & 7 & 11.63 & 80.00 & \textbf{92.00} \\
medium & spatial & task & 7 & 11.63 & 84.00 & \textbf{96.00} \\
medium & goal & object & 4 & 10.71 & 92.00 & \textbf{100.00} \\
medium & goal & object & 0 & 10.20 & 96.00 & \textbf{100.00} \\
medium & spatial & task & 2 & 7.84 & 92.00 & \textbf{100.00} \\
medium & 10 & task & 0 & 6.12 & 84.00 & \textbf{92.00} \\
medium & 10 & task & 8 & 5.00 & 80.00 & \textbf{96.00} \\
medium & goal & object & 1 & 4.08 & 100.00 & \textbf{100.00} \\
medium & object & object & 0 & 2.04 & 76.00 & \textbf{100.00} \\
medium & 10 & object & 5 & 1.96 & 96.00 & \textbf{100.00} \\
medium & goal & task & 5 & 1.96 & 92.00 & \textbf{92.00} \\
\midrule
hard & 10 & object & 0 & 0.00 & 0.00 & \textbf{0.00} \\
hard & 10 & object & 6 & 0.00 & 92.00 & \textbf{100.00} \\
hard & 10 & object & 7 & 0.00 & 0.00 & \textbf{0.00} \\
hard & 10 & swap & 3 & 0.00 & 0.00 & \textbf{0.00} \\
hard & 10 & swap & 8 & 0.00 & 0.00 & \textbf{0.00} \\
hard & 10 & swap & 9 & 0.00 & 0.00 & \textbf{0.00} \\
hard & 10 & task & 1 & 0.00 & 0.00 & \textbf{0.00} \\
hard & 10 & task & 2 & 0.00 & 0.00 & \textbf{0.00} \\
hard & 10 & task & 3 & 0.00 & 0.00 & \textbf{0.00} \\
hard & 10 & task & 4 & 0.00 & 0.00 & \textbf{0.00} \\
hard & 10 & task & 5 & 0.00 & 0.00 & \textbf{0.00} \\
hard & 10 & task & 6 & 0.00 & 0.00 & \textbf{0.00} \\
hard & 10 & task & 7 & 0.00 & 0.00 & \textbf{0.00} \\
hard & 10 & task & 9 & 0.00 & 0.00 & \textbf{0.00} \\
hard & goal & object & 3 & 0.00 & 0.00 & \textbf{0.00} \\
hard & goal & object & 6 & 0.00 & 36.00 & \textbf{48.00} \\
hard & goal & object & 8 & 0.00 & 80.00 & \textbf{92.00} \\
hard & goal & swap & 0 & 0.00 & 0.00 & \textbf{0.00} \\
hard & goal & swap & 1 & 0.00 & 0.00 & \textbf{0.00} \\
hard & goal & swap & 2 & 0.00 & 0.00 & \textbf{96.00} \\
hard & goal & swap & 3 & 0.00 & 88.00 & \textbf{88.00} \\
hard & goal & swap & 4 & 0.00 & 0.00 & \textbf{0.00} \\
hard & goal & swap & 6 & 0.00 & 0.00 & \textbf{0.00} \\
hard & goal & swap & 7 & 0.00 & 0.00 & \textbf{100.00} \\
hard & goal & swap & 9 & 0.00 & 0.00 & \textbf{100.00} \\
hard & goal & task & 0 & 0.00 & 0.00 & \textbf{100.00} \\
hard & goal & task & 1 & 0.00 & 0.00 & \textbf{0.00} \\
hard & goal & task & 2 & 0.00 & 0.00 & \textbf{0.00} \\
hard & goal & task & 3 & 0.00 & 0.00 & \textbf{0.00} \\
hard & goal & task & 4 & 0.00 & 0.00 & \textbf{0.00} \\
hard & goal & task & 6 & 0.00 & 0.00 & \textbf{0.00} \\
hard & goal & task & 8 & 0.00 & 4.00 & \textbf{0.00} \\
hard & goal & task & 9 & 0.00 & 0.00 & \textbf{0.00} \\
hard & object & task & 0 & 0.00 & 0.00 & \textbf{0.00} \\
hard & object & task & 2 & 0.00 & 0.00 & \textbf{0.00} \\
hard & object & task & 3 & 0.00 & 0.00 & \textbf{0.00} \\
hard & object & task & 4 & 0.00 & 0.00 & \textbf{0.00} \\
hard & object & task & 5 & 0.00 & 0.00 & \textbf{0.00} \\
hard & object & task & 6 & 0.00 & 0.00 & \textbf{0.00} \\
hard & object & task & 7 & 0.00 & 0.00 & \textbf{0.00} \\
hard & object & task & 8 & 0.00 & 0.00 & \textbf{0.00} \\
hard & object & task & 9 & 0.00 & 0.00 & \textbf{0.00} \\
hard & spatial & swap & 3 & 0.00 & 0.00 & \textbf{0.00} \\
hard & spatial & task & 0 & 0.00 & 0.00 & \textbf{80.00} \\
hard & spatial & task & 1 & 0.00 & 0.00 & \textbf{0.00} \\
hard & spatial & task & 3 & 0.00 & 92.00 & \textbf{0.00} \\
hard & spatial & task & 4 & 0.00 & 56.00 & \textbf{0.00} \\
hard & spatial & task & 6 & 0.00 & 96.00 & \textbf{100.00} \\
hard & spatial & task & 8 & 0.00 & 80.00 & \textbf{92.00} \\
\end{longtable}
\begin{longtable}{@{}llllccc@{}}
\caption{Per-task success rate (\%) on LIBERO-PRO with $\pi_0$, sorted by difficulty bucket and PPO init SR.\label{tab:libero-pertask-pi0}}\\
\toprule
\textbf{Diff.} & \textbf{Suite} & \textbf{Pert.} & \textbf{Task} & \textbf{Init} & \textbf{PPO} & \textbf{Ours} \\
\midrule
\endfirsthead
\multicolumn{7}{c}{\textit{(Table~\ref{tab:libero-pertask-pi0} continued)}}\\
\toprule
\textbf{Diff.} & \textbf{Suite} & \textbf{Pert.} & \textbf{Task} & \textbf{Init} & \textbf{PPO} & \textbf{EMA (Ours)} \\
\midrule
\endhead
\midrule \multicolumn{7}{r}{\textit{(continued on next page)}} \\
\endfoot
\bottomrule
\endlastfoot
easy & goal & object & 1 & 93.88 & 100.00 & \textbf{100.00} \\
easy & object & swap & 4 & 89.29 & 96.00 & \textbf{100.00} \\
easy & goal & swap & 8 & 86.67 & 100.00 & \textbf{100.00} \\
easy & spatial & object & 2 & 86.27 & 100.00 & \textbf{100.00} \\
easy & spatial & swap & 2 & 82.35 & 100.00 & \textbf{100.00} \\
easy & spatial & object & 0 & 77.55 & 8.00 & \textbf{96.00} \\
easy & spatial & swap & 0 & 77.55 & 92.00 & \textbf{88.00} \\
easy & goal & object & 5 & 74.51 & 0.00 & \textbf{96.00} \\
easy & spatial & object & 3 & 66.67 & 96.00 & \textbf{100.00} \\
easy & object & object & 4 & 62.50 & 100.00 & \textbf{100.00} \\
easy & object & swap & 8 & 60.00 & 100.00 & \textbf{100.00} \\
easy & spatial & task & 5 & 56.86 & 100.00 & \textbf{96.00} \\
easy & object & swap & 9 & 56.36 & 92.00 & \textbf{92.00} \\
easy & object & swap & 0 & 55.10 & 100.00 & \textbf{100.00} \\
easy & object & swap & 2 & 54.90 & 96.00 & \textbf{96.00} \\
easy & goal & object & 8 & 53.33 & 80.00 & \textbf{0.00} \\
easy & goal & swap & 5 & 50.98 & 100.00 & \textbf{96.00} \\
easy & object & swap & 5 & 47.06 & 96.00 & \textbf{96.00} \\
easy & goal & object & 7 & 46.51 & 100.00 & \textbf{100.00} \\
easy & spatial & object & 4 & 44.64 & 80.00 & \textbf{12.00} \\
easy & goal & task & 7 & 44.19 & 68.00 & \textbf{92.00} \\
easy & object & swap & 6 & 43.40 & 100.00 & \textbf{100.00} \\
easy & object & object & 5 & 43.14 & 96.00 & \textbf{92.00} \\
easy & spatial & swap & 4 & 41.07 & 88.00 & \textbf{64.00} \\
easy & object & object & 1 & 40.82 & 100.00 & \textbf{100.00} \\
easy & object & swap & 1 & 40.82 & 100.00 & \textbf{96.00} \\
easy & object & object & 8 & 40.00 & 100.00 & \textbf{100.00} \\
\midrule
medium & object & object & 6 & 39.62 & 96.00 & \textbf{100.00} \\
medium & spatial & task & 1 & 32.65 & 88.00 & \textbf{20.00} \\
medium & goal & object & 9 & 27.27 & 44.00 & \textbf{72.00} \\
medium & spatial & swap & 8 & 16.67 & 72.00 & \textbf{88.00} \\
medium & spatial & swap & 1 & 16.33 & 100.00 & \textbf{92.00} \\
medium & spatial & object & 1 & 14.29 & 100.00 & \textbf{100.00} \\
medium & object & swap & 7 & 13.95 & 96.00 & \textbf{92.00} \\
medium & spatial & object & 8 & 13.33 & 88.00 & \textbf{88.00} \\
medium & object & object & 9 & 12.73 & 96.00 & \textbf{96.00} \\
medium & object & object & 0 & 12.24 & 100.00 & \textbf{100.00} \\
medium & spatial & swap & 6 & 11.32 & 100.00 & \textbf{76.00} \\
medium & goal & swap & 3 & 11.11 & 100.00 & \textbf{96.00} \\
medium & spatial & object & 9 & 10.91 & 96.00 & \textbf{92.00} \\
medium & object & object & 7 & 9.30 & 100.00 & \textbf{100.00} \\
medium & goal & swap & 7 & 6.98 & 0.00 & \textbf{96.00} \\
medium & object & swap & 3 & 6.67 & 92.00 & \textbf{100.00} \\
medium & spatial & task & 8 & 6.67 & 92.00 & \textbf{100.00} \\
medium & spatial & task & 2 & 5.88 & 88.00 & \textbf{84.00} \\
medium & spatial & object & 6 & 5.66 & 68.00 & \textbf{96.00} \\
medium & spatial & swap & 9 & 5.45 & 88.00 & \textbf{84.00} \\
medium & goal & object & 3 & 4.44 & 16.00 & \textbf{72.00} \\
medium & goal & task & 5 & 3.92 & 52.00 & \textbf{80.00} \\
medium & object & object & 2 & 3.92 & 0.00 & \textbf{100.00} \\
medium & 10 & object & 1 & 2.04 & 0.00 & \textbf{16.00} \\
\midrule
hard & 10 & object & 0 & 0.00 & 0.00 & \textbf{0.00} \\
hard & 10 & object & 2 & 0.00 & 0.00 & \textbf{0.00} \\
hard & 10 & object & 3 & 0.00 & 88.00 & \textbf{84.00} \\
hard & 10 & object & 4 & 0.00 & 0.00 & \textbf{16.00} \\
hard & 10 & object & 5 & 0.00 & 0.00 & \textbf{0.00} \\
hard & 10 & object & 6 & 0.00 & 0.00 & \textbf{0.00} \\
hard & 10 & object & 7 & 0.00 & 0.00 & \textbf{0.00} \\
hard & 10 & object & 8 & 0.00 & 0.00 & \textbf{0.00} \\
hard & 10 & object & 9 & 0.00 & 0.00 & \textbf{0.00} \\
hard & 10 & swap & 0 & 0.00 & 0.00 & \textbf{0.00} \\
hard & 10 & swap & 1 & 0.00 & 0.00 & \textbf{56.00} \\
hard & 10 & swap & 2 & 0.00 & 20.00 & \textbf{0.00} \\
hard & 10 & swap & 3 & 0.00 & 0.00 & \textbf{0.00} \\
hard & 10 & swap & 4 & 0.00 & 12.00 & \textbf{0.00} \\
hard & 10 & swap & 5 & 0.00 & 0.00 & \textbf{0.00} \\
hard & 10 & swap & 6 & 0.00 & 0.00 & \textbf{0.00} \\
hard & 10 & swap & 7 & 0.00 & 0.00 & \textbf{0.00} \\
hard & 10 & swap & 8 & 0.00 & 0.00 & \textbf{0.00} \\
hard & 10 & swap & 9 & 0.00 & 0.00 & \textbf{0.00} \\
hard & goal & object & 0 & 0.00 & 20.00 & \textbf{0.00} \\
hard & goal & object & 2 & 0.00 & 0.00 & \textbf{0.00} \\
hard & goal & object & 4 & 0.00 & 24.00 & \textbf{0.00} \\
hard & goal & object & 6 & 0.00 & 0.00 & \textbf{0.00} \\
hard & goal & swap & 0 & 0.00 & 0.00 & \textbf{8.00} \\
hard & goal & swap & 1 & 0.00 & 100.00 & \textbf{0.00} \\
hard & goal & swap & 2 & 0.00 & 0.00 & \textbf{0.00} \\
hard & goal & swap & 4 & 0.00 & 0.00 & \textbf{0.00} \\
hard & goal & swap & 6 & 0.00 & 0.00 & \textbf{0.00} \\
hard & goal & swap & 9 & 0.00 & 0.00 & \textbf{0.00} \\
hard & goal & task & 0 & 0.00 & 0.00 & \textbf{0.00} \\
hard & goal & task & 1 & 0.00 & 100.00 & \textbf{0.00} \\
hard & goal & task & 2 & 0.00 & 0.00 & \textbf{0.00} \\
hard & goal & task & 3 & 0.00 & 0.00 & \textbf{0.00} \\
hard & goal & task & 4 & 0.00 & 0.00 & \textbf{0.00} \\
hard & goal & task & 6 & 0.00 & 0.00 & \textbf{0.00} \\
hard & goal & task & 8 & 0.00 & 0.00 & \textbf{0.00} \\
hard & goal & task & 9 & 0.00 & 0.00 & \textbf{0.00} \\
hard & object & object & 3 & 0.00 & 92.00 & \textbf{96.00} \\
hard & spatial & object & 5 & 0.00 & 0.00 & \textbf{40.00} \\
hard & spatial & object & 7 & 0.00 & 0.00 & \textbf{0.00} \\
hard & spatial & swap & 3 & 0.00 & 4.00 & \textbf{0.00} \\
hard & spatial & swap & 5 & 0.00 & 52.00 & \textbf{20.00} \\
hard & spatial & swap & 7 & 0.00 & 4.00 & \textbf{12.00} \\
hard & spatial & task & 0 & 0.00 & 0.00 & \textbf{0.00} \\
hard & spatial & task & 3 & 0.00 & 0.00 & \textbf{72.00} \\
hard & spatial & task & 4 & 0.00 & 76.00 & \textbf{20.00} \\
hard & spatial & task & 6 & 0.00 & 0.00 & \textbf{40.00} \\
hard & spatial & task & 7 & 0.00 & 4.00 & \textbf{92.00} \\
hard & spatial & task & 9 & 0.00 & 0.00 & \textbf{68.00} \\
\end{longtable}

\begin{longtable}{@{}llllccc@{}}
\caption{Per-task success rate (\%) on ManiSkill at $\tau{=}1.0$, sorted by difficulty bucket and SFT init SR.\label{tab:maniskill-pertask}}\\
\toprule
\textbf{Difficulty} & \textbf{Category} & \textbf{Variant} & \textbf{Task} & \textbf{SFT} & \textbf{PPO} & \textbf{Ours} \\
\midrule
\endfirsthead
\multicolumn{7}{c}{\textit{(Table~\ref{tab:maniskill-pertask} continued)}}\\
\toprule
\textbf{Difficulty} & \textbf{Category} & \textbf{Variant} & \textbf{Task} & \textbf{SFT} & \textbf{PPO} & \textbf{Ours} \\
\midrule
\endhead
\midrule \multicolumn{7}{r}{\textit{(continued on next page)}} \\
\endfoot
\bottomrule
\endlastfoot
hard & semantic & MultiPlate & 0 & 23.10 & 30.80 & \textbf{33.30} \\
hard & semantic & MultiPlate & 13 & 23.10 & 30.80 & \textbf{25.00} \\
hard & semantic & MultiPlate & 11 & 22.20 & 33.30 & \textbf{66.70} \\
hard & execution & Position & 1 & 21.40 & 57.10 & \textbf{71.40} \\
hard & semantic & MultiPlate & 7 & 20.00 & 20.00 & \textbf{45.00} \\
hard & execution & PositionChangeTo & 8 & 20.00 & 55.00 & \textbf{60.00} \\
hard & execution & EEPose & 3 & 20.00 & 70.00 & \textbf{60.00} \\
hard & execution & EEPose & 5 & 20.00 & 73.30 & \textbf{93.30} \\
hard & visual & VisionTexture05 & 11 & 20.00 & 53.30 & \textbf{100.00} \\
hard & visual & VisionWhole05 & 11 & 20.00 & 53.30 & \textbf{100.00} \\
hard & visual & VisionWhole05 & 14 & 20.00 & 60.00 & \textbf{70.00} \\
hard & execution & Position & 2 & 18.80 & 56.20 & \textbf{68.80} \\
hard & execution & PositionChangeTo & 4 & 18.80 & 68.80 & \textbf{62.50} \\
hard & semantic & MultiPlate & 9 & 18.20 & 45.50 & \textbf{36.40} \\
hard & visual & VisionWhole03 & 6 & 18.20 & 100.00 & \textbf{84.60} \\
hard & semantic & MultiCarrot & 24 & 17.90 & 32.10 & \textbf{35.70} \\
hard & execution & PositionChangeTo & 12 & 17.60 & 52.90 & \textbf{85.70} \\
hard & execution & Position & 12 & 16.70 & 61.10 & \textbf{61.10} \\
hard & execution & PositionChangeTo & 14 & 16.70 & 75.00 & \textbf{58.30} \\
hard & semantic & MultiCarrot & 18 & 15.80 & 52.60 & \textbf{57.90} \\
hard & execution & Position & 11 & 15.40 & 30.80 & \textbf{38.50} \\
hard & visual & VisionWhole03 & 11 & 15.40 & 53.30 & \textbf{100.00} \\
hard & semantic & MultiPlate & 5 & 15.00 & 65.00 & \textbf{60.00} \\
hard & semantic & Plate & 2 & 14.30 & 81.00 & \textbf{85.70} \\
hard & semantic & MultiPlate & 3 & 14.30 & 64.30 & \textbf{83.30} \\
hard & semantic & MultiPlate & 8 & 14.30 & 42.90 & \textbf{50.00} \\
hard & execution & PositionChangeTo & 7 & 14.30 & 42.90 & \textbf{42.90} \\
hard & visual & VisionTexture05 & 2 & 14.30 & 76.20 & \textbf{81.00} \\
hard & visual & VisionWhole05 & 2 & 14.30 & 66.70 & \textbf{61.90} \\
hard & execution & PositionChangeTo & 10 & 13.30 & 40.00 & \textbf{50.00} \\
hard & execution & PositionChangeTo & 1 & 12.50 & 62.50 & \textbf{50.00} \\
hard & visual & VisionTexture03 & 12 & 12.50 & 81.20 & \textbf{93.80} \\
hard & semantic & MultiCarrot & 21 & 11.80 & 26.50 & \textbf{44.10} \\
hard & semantic & MultiPlate & 6 & 11.80 & 29.40 & \textbf{40.00} \\
hard & semantic & MainCarrot & 21 & 9.70 & 77.40 & \textbf{67.70} \\
hard & semantic & MultiCarrot & 23 & 9.70 & 35.50 & \textbf{44.40} \\
hard & visual & VisionTexture05 & 1 & 9.50 & 42.90 & \textbf{66.70} \\
hard & visual & VisionWhole05 & 1 & 9.50 & 38.10 & \textbf{50.00} \\
hard & semantic & MultiPlate & 4 & 9.10 & 45.50 & \textbf{36.40} \\
hard & execution & EEPose & 1 & 8.30 & 50.00 & \textbf{100.00} \\
hard & execution & PositionChangeTo & 2 & 7.70 & 53.80 & \textbf{100.00} \\
hard & execution & PositionChangeTo & 6 & 7.70 & 76.90 & \textbf{100.00} \\
hard & execution & EEPose & 11 & 7.10 & 64.30 & \textbf{50.00} \\
hard & semantic & MultiPlate & 2 & 6.20 & 31.20 & \textbf{33.30} \\
hard & semantic & MultiPlate & 14 & 6.20 & 50.00 & \textbf{50.00} \\
hard & semantic & MainCarrot & 22 & 5.90 & 47.10 & \textbf{61.80} \\
hard & semantic & MultiPlate & 10 & 5.90 & 5.90 & \textbf{17.60} \\
hard & execution & PositionChangeTo & 3 & 5.90 & 35.30 & \textbf{47.10} \\
hard & execution & PositionChangeTo & 9 & 5.90 & 47.10 & \textbf{50.00} \\
hard & visual & VisionWhole03 & 1 & 5.00 & 61.90 & \textbf{66.70} \\
hard & semantic & Plate & 1 & 4.80 & 52.40 & \textbf{57.10} \\
hard & visual & VisionTexture03 & 1 & 4.80 & 38.10 & \textbf{66.70} \\
hard & visual & VisionTexture03 & 2 & 4.80 & 100.00 & \textbf{100.00} \\
hard & semantic & MultiCarrot & 22 & 2.60 & 59.00 & \textbf{64.30} \\
hard & semantic & MultiPlate & 1 & 0.00 & 28.60 & \textbf{40.00} \\
hard & semantic & MultiPlate & 12 & 0.00 & 33.30 & \textbf{41.70} \\
hard & semantic & MultiPlate & 15 & 0.00 & 44.40 & \textbf{33.30} \\
hard & execution & PositionChangeTo & 11 & 0.00 & 75.00 & \textbf{33.30} \\
hard & execution & PositionChangeTo & 13 & 0.00 & 10.00 & \textbf{33.30} \\
\midrule
medium & semantic & MainCarrot & 18 & 50.00 & 68.80 & \textbf{81.20} \\
medium & semantic & Plate & 14 & 50.00 & 80.00 & \textbf{100.00} \\
medium & semantic & MainCarrot & 19 & 48.60 & 68.60 & \textbf{77.10} \\
medium & semantic & MainCarrot & 16 & 48.30 & 75.90 & \textbf{89.70} \\
medium & execution & Position & 5 & 47.80 & 73.90 & \textbf{87.00} \\
medium & visual & VisionTexture05 & 7 & 47.10 & 76.50 & \textbf{76.50} \\
medium & visual & VisionWhole05 & 3 & 47.10 & 76.50 & \textbf{100.00} \\
medium & execution & EEPose & 10 & 46.70 & 60.00 & \textbf{73.30} \\
medium & visual & VisionImage & 14 & 46.20 & 92.30 & \textbf{100.00} \\
medium & visual & VisionImage & 11 & 44.40 & 66.70 & \textbf{100.00} \\
medium & semantic & Plate & 12 & 43.80 & 68.80 & \textbf{100.00} \\
medium & execution & Position & 9 & 43.80 & 25.00 & \textbf{62.50} \\
medium & execution & EEPose & 7 & 43.80 & 37.50 & \textbf{50.00} \\
medium & execution & EEPose & 9 & 42.90 & 42.90 & \textbf{71.40} \\
medium & visual & VisionWhole05 & 5 & 42.90 & 78.60 & \textbf{78.60} \\
medium & visual & VisionWhole05 & 9 & 42.10 & 57.90 & \textbf{73.70} \\
medium & visual & VisionTexture05 & 15 & 41.20 & 52.90 & \textbf{58.80} \\
medium & visual & VisionWhole03 & 15 & 41.20 & 52.90 & \textbf{64.70} \\
medium & execution & Position & 8 & 40.00 & 66.70 & \textbf{73.30} \\
medium & visual & VisionImage & 12 & 40.00 & 85.00 & \textbf{95.00} \\
medium & visual & VisionTexture05 & 14 & 40.00 & 80.00 & \textbf{100.00} \\
medium & visual & VisionWhole03 & 14 & 40.00 & 90.00 & \textbf{90.00} \\
medium & execution & PositionChangeTo & 0 & 38.10 & 57.10 & \textbf{71.40} \\
medium & semantic & MainCarrot & 20 & 37.50 & 95.80 & \textbf{100.00} \\
medium & visual & VisionWhole05 & 12 & 37.50 & 68.80 & \textbf{81.20} \\
medium & execution & EEPose & 14 & 36.80 & 73.70 & \textbf{84.20} \\
medium & visual & VisionWhole03 & 8 & 35.30 & 88.20 & \textbf{88.20} \\
medium & execution & Position & 14 & 35.00 & 65.00 & \textbf{75.00} \\
medium & semantic & Plate & 11 & 33.30 & 66.70 & \textbf{73.30} \\
medium & execution & Position & 3 & 33.30 & 53.30 & \textbf{46.70} \\
medium & execution & EEPose & 6 & 33.30 & 77.80 & \textbf{77.80} \\
medium & visual & VisionTexture03 & 11 & 33.30 & 73.30 & \textbf{100.00} \\
medium & visual & VisionWhole05 & 10 & 33.30 & 46.70 & \textbf{40.00} \\
medium & execution & EEPose & 15 & 31.60 & 21.10 & \textbf{31.60} \\
medium & semantic & Plate & 6 & 30.80 & 69.20 & \textbf{84.60} \\
medium & visual & VisionTexture03 & 6 & 30.80 & 100.00 & \textbf{100.00} \\
medium & visual & VisionWhole05 & 6 & 30.80 & 53.80 & \textbf{100.00} \\
medium & semantic & MainCarrot & 23 & 30.40 & 34.80 & \textbf{56.50} \\
medium & execution & EEPose & 4 & 29.40 & 76.50 & \textbf{88.20} \\
medium & execution & EEPose & 12 & 29.40 & 88.20 & \textbf{100.00} \\
medium & visual & VisionWhole05 & 15 & 29.40 & 29.40 & \textbf{47.10} \\
medium & semantic & MultiCarrot & 19 & 29.20 & 33.30 & \textbf{37.50} \\
medium & semantic & MultiCarrot & 16 & 28.60 & 35.70 & \textbf{54.50} \\
medium & execution & EEPose & 8 & 28.60 & 66.70 & \textbf{71.40} \\
medium & visual & VisionTexture03 & 5 & 28.60 & 78.60 & \textbf{100.00} \\
medium & visual & VisionTexture05 & 5 & 28.60 & 57.10 & \textbf{75.00} \\
medium & visual & VisionWhole03 & 2 & 27.80 & 90.50 & \textbf{81.00} \\
medium & execution & Position & 15 & 27.30 & 27.30 & \textbf{36.40} \\
medium & visual & VisionImage & 1 & 26.70 & 33.30 & \textbf{66.70} \\
medium & semantic & MainCarrot & 24 & 26.10 & 65.20 & \textbf{73.90} \\
medium & semantic & MultiCarrot & 17 & 25.00 & 50.00 & \textbf{57.10} \\
medium & execution & PositionChangeTo & 5 & 25.00 & 75.00 & \textbf{87.50} \\
medium & visual & VisionTexture05 & 12 & 25.00 & 75.00 & \textbf{81.20} \\
medium & semantic & MultiCarrot & 20 & 24.20 & 66.70 & \textbf{75.80} \\
medium & execution & Position & 6 & 23.50 & 47.10 & \textbf{88.20} \\
medium & execution & Position & 13 & 23.50 & 35.30 & \textbf{58.80} \\
medium & visual & VisionTexture05 & 3 & 23.50 & 82.40 & \textbf{100.00} \\
medium & execution & PositionChangeTo & 15 & 23.10 & 34.60 & \textbf{30.80} \\
medium & visual & VisionTexture05 & 6 & 23.10 & 69.20 & \textbf{100.00} \\
\midrule
easy & visual & VisionImage & 13 & 92.30 & 61.50 & \textbf{76.90} \\
easy & visual & VisionTexture03 & 0 & 88.90 & 100.00 & \textbf{100.00} \\
easy & visual & VisionWhole03 & 0 & 88.20 & 88.90 & \textbf{100.00} \\
easy & visual & VisionImage & 7 & 87.00 & 82.60 & \textbf{95.70} \\
easy & visual & VisionImage & 8 & 86.70 & 86.70 & \textbf{100.00} \\
easy & visual & VisionImage & 0 & 84.60 & 84.60 & \textbf{96.20} \\
easy & semantic & Plate & 0 & 83.30 & 88.90 & \textbf{94.40} \\
easy & visual & VisionTexture05 & 0 & 83.30 & 83.30 & \textbf{75.00} \\
easy & visual & VisionWhole03 & 10 & 76.90 & 66.70 & \textbf{80.00} \\
easy & visual & VisionTexture03 & 3 & 76.50 & 94.10 & \textbf{100.00} \\
easy & visual & VisionImage & 10 & 75.00 & 58.30 & \textbf{66.70} \\
easy & visual & VisionWhole03 & 9 & 75.00 & 78.90 & \textbf{73.70} \\
easy & semantic & Plate & 10 & 73.30 & 33.30 & \textbf{46.70} \\
easy & execution & Position & 4 & 73.30 & 20.00 & \textbf{73.30} \\
easy & visual & VisionImage & 15 & 73.30 & 46.70 & \textbf{80.00} \\
easy & visual & VisionImage & 5 & 71.40 & 100.00 & \textbf{100.00} \\
easy & visual & VisionImage & 4 & 70.60 & 94.10 & \textbf{100.00} \\
easy & visual & VisionTexture03 & 7 & 70.60 & 76.50 & \textbf{100.00} \\
easy & visual & VisionImage & 3 & 70.00 & 90.00 & \textbf{100.00} \\
easy & visual & VisionTexture03 & 14 & 70.00 & 90.00 & \textbf{100.00} \\
easy & semantic & Plate & 9 & 68.40 & 57.90 & \textbf{68.40} \\
easy & visual & VisionTexture03 & 10 & 66.70 & 40.00 & \textbf{80.00} \\
easy & visual & VisionWhole05 & 0 & 66.70 & 72.20 & \textbf{75.00} \\
easy & semantic & Plate & 3 & 64.70 & 94.10 & \textbf{94.10} \\
easy & semantic & Plate & 7 & 64.70 & 70.60 & \textbf{88.20} \\
easy & visual & VisionTexture03 & 8 & 64.70 & 100.00 & \textbf{100.00} \\
easy & visual & VisionTexture05 & 8 & 64.70 & 94.10 & \textbf{100.00} \\
easy & visual & VisionWhole05 & 8 & 64.70 & 70.60 & \textbf{83.30} \\
easy & execution & Position & 0 & 64.30 & 50.00 & \textbf{57.10} \\
easy & visual & VisionImage & 9 & 64.30 & 50.00 & \textbf{100.00} \\
easy & visual & VisionTexture03 & 9 & 63.20 & 47.40 & \textbf{80.00} \\
easy & visual & VisionTexture05 & 9 & 63.20 & 68.40 & \textbf{73.70} \\
easy & visual & VisionWhole03 & 7 & 62.50 & 82.40 & \textbf{100.00} \\
easy & visual & VisionWhole03 & 12 & 62.50 & 93.80 & \textbf{93.80} \\
easy & semantic & Plate & 13 & 60.00 & 50.00 & \textbf{80.00} \\
easy & visual & VisionTexture05 & 13 & 60.00 & 40.00 & \textbf{80.00} \\
easy & visual & VisionWhole03 & 13 & 60.00 & 70.00 & \textbf{60.00} \\
easy & semantic & Plate & 8 & 58.80 & 94.10 & \textbf{94.10} \\
easy & execution & Position & 7 & 58.30 & 41.70 & \textbf{75.00} \\
easy & execution & EEPose & 13 & 58.30 & 41.70 & \textbf{66.70} \\
easy & semantic & Plate & 5 & 57.10 & 100.00 & \textbf{100.00} \\
easy & visual & VisionImage & 6 & 57.10 & 85.70 & \textbf{92.90} \\
easy & visual & VisionWhole03 & 5 & 57.10 & 100.00 & \textbf{100.00} \\
easy & semantic & Plate & 4 & 56.20 & 93.80 & \textbf{93.80} \\
easy & visual & VisionTexture03 & 4 & 56.20 & 100.00 & \textbf{100.00} \\
easy & visual & VisionWhole05 & 4 & 56.20 & 81.20 & \textbf{100.00} \\
easy & execution & EEPose & 0 & 53.80 & 84.60 & \textbf{92.30} \\
easy & visual & VisionTexture05 & 10 & 53.30 & 26.70 & \textbf{53.30} \\
easy & visual & VisionWhole03 & 4 & 53.30 & 93.80 & \textbf{100.00} \\
easy & semantic & Plate & 15 & 52.90 & 41.20 & \textbf{58.80} \\
easy & visual & VisionImage & 2 & 52.90 & 88.20 & \textbf{88.20} \\
easy & visual & VisionTexture03 & 15 & 52.90 & 52.90 & \textbf{41.20} \\
easy & visual & VisionWhole03 & 3 & 52.90 & 94.10 & \textbf{100.00} \\
easy & visual & VisionWhole05 & 7 & 52.90 & 70.60 & \textbf{88.20} \\
easy & semantic & MainCarrot & 17 & 52.00 & 64.00 & \textbf{92.00} \\
easy & execution & Position & 10 & 50.00 & 25.00 & \textbf{50.00} \\
easy & execution & EEPose & 2 & 50.00 & 64.30 & \textbf{64.30} \\
easy & visual & VisionTexture03 & 13 & 50.00 & 50.00 & \textbf{90.00} \\
easy & visual & VisionTexture05 & 4 & 50.00 & 93.80 & \textbf{100.00} \\
easy & visual & VisionWhole05 & 13 & 50.00 & 40.00 & \textbf{60.00} \\
\end{longtable}
\section{LLM Experiment Details}
\label{app:sec:llm}

This appendix provides implementation details for the language-model experiments in
Section~\ref{subsec:exp:llm}. We use the same PDE interface as in the VLA
experiments: each problem $g$ has a canonical prompt $p_g$, PDE proposes an
alternative prompt $p \sim \rho(\cdot \mid g,\mathcal{H})$ for training, and
evaluation is always performed under $p_g$.

\subsection{Experiment: AIME 2026}
\label{app:aime}

\textbf{Setup.}
We evaluate on AIME 2026, which contains $30$ olympiad-style math problems, each
with a unique answer given as a three-digit integer between $000$ and $999$. We
fine-tune Qwen3-4B with GRPO, using Claude Sonnet 4.6 as the prompt sampler
$\rho$. Training runs for $T=3$ PDE iterations. At each iteration, we sample
$K=1$ alternative prompt per problem, generate $N=8$ rollouts per prompt, and run
$10$ epochs of GRPO. Both GRPO and GRPO+PDE use learning rate
$3 \cdot 10^{-6}$.

\textbf{Implementation details.}
For the first PDE iteration, we query the prompt sampler $\rho$ once per problem
$g$ to obtain an alternative prompt $p$. In later iterations, we reuse $p$ if it
outperforms the canonical prompt $p_g$ after the previous training iteration.
Otherwise, we query $\rho$ for a new prompt, conditioning on the canonical
prompt, the previous alternative prompt, and the rollout history
$\mathcal{H}$. Thus, prompt admission is determined adaptively by comparison to
the canonical prompt rather than by the positive-success criterion used in the VLA experiments.

For mixed backpropagation, this experiment uses an arithmetic-mixture variant:
\[
\log \pi_{\mathrm{mix}}(a \mid o)
=
\log\!\left(
\pi_\theta(a \mid o,p_g) + \pi_\theta(a \mid o,p)
\right),
\]
up to an additive constant. This differs from the geometric-mean form used in
the VLA experiments.

\begin{figure}[h!]
\centering
\includegraphics[width=0.8\columnwidth]{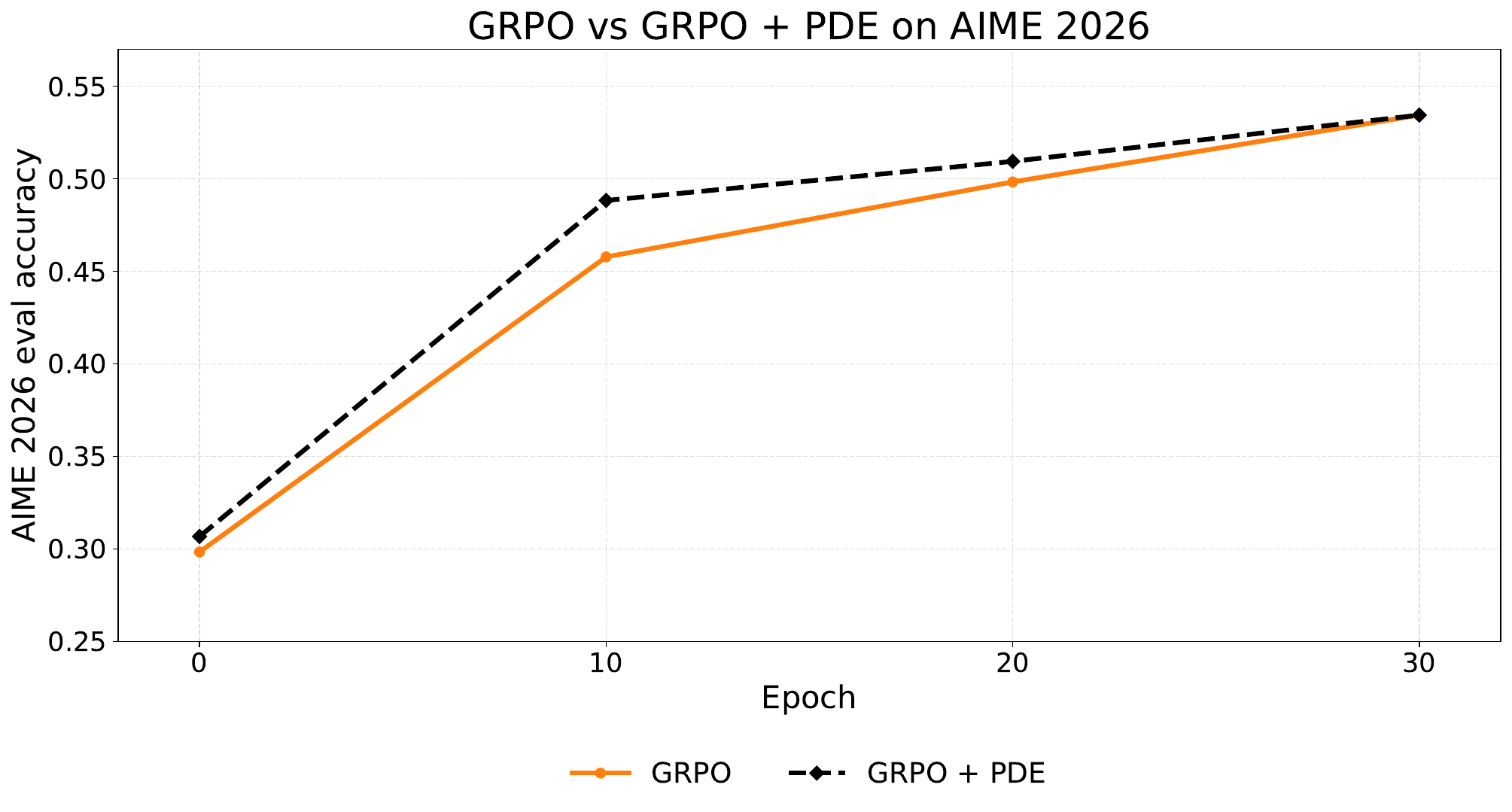}
\caption{AIME 2026 accuracy for GRPO and GRPO+PDE. Evaluation uses only the
canonical problem prompt $p_g$.}
\label{fig:app_aime_pde}
\end{figure}

\textbf{Evaluation and results.}
At each checkpoint, we evaluate under the canonical prompt $p_g$ only. For each
problem, we generate $20$ independent rollouts and report accuracy over all
$30 \times 20 = 600$ generations. Figure~\ref{fig:app_aime_pde} compares GRPO
and GRPO+PDE. GRPO+PDE improves early learning, reaching $48.9\%$ accuracy after
the first iteration compared with $45.8\%$ for GRPO, and both methods reach
$53.3\%$ after three iterations.

\subsection{Experiment: LiveCodeBench}
\label{app:lcb}

\textbf{Setup.}
We evaluate on LiveCodeBench, which contains $1{,}055$ competitive-programming
problems with hidden tests. We use $600$ problems for training and reserve the
remaining $455$ for held-out evaluation. We fine-tune with RLOO at learning rate
$5 \cdot 10^{-5}$.

\textbf{Implementation details.}
For each eligible training problem $g$, the policy itself serves as the prompt
sampler $\rho$. The policy first generates $10$ candidate solutions under the
canonical prompt $p_g$. It is then re-prompted to produce an alternative prompt
$p$ conditioned on $p_g$ and its own raw rollouts. The sampler observes the
rollouts but not their rewards. This refinement loop is run for $3$ rounds, and
the best alternative prompt is cached. During training, the cached prompt is used
alongside the canonical prompt in the RLOO update. The first $50\%$ of training
problems are marked as eligible for prompt refinement; held-out evaluation uses
only the canonical prompt $p_g$.

\begin{figure}[h!]
\centering
\includegraphics[width=0.8\columnwidth]{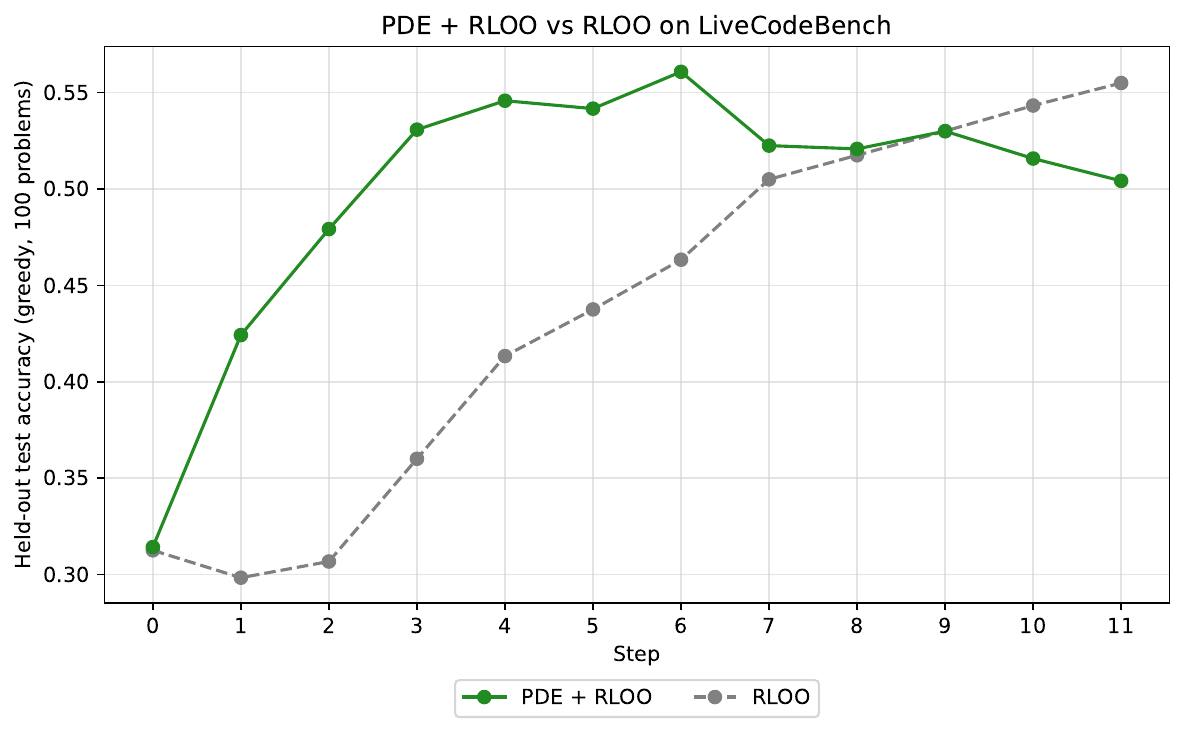}
\caption{LiveCodeBench held-out accuracy for RLOO and PDE+RLOO. Training uses
alternative prompts from PDE, while evaluation uses only the canonical prompt
$p_g$.}
\label{fig:app_lcb_pde}
\end{figure}

\textbf{Evaluation and results.}
At each batch checkpoint, we evaluate on the first $100$ held-out problems by
generating one greedy rollout per problem and reporting the fraction of solutions
that pass all hidden tests. The final checkpoint is evaluated in the same way on
$400$ held-out problems. Figure~\ref{fig:app_lcb_pde} compares RLOO and
PDE+RLOO. PDE+RLOO learns substantially faster: it reaches $50\%$ held-out
accuracy by batch $4$, while RLOO reaches this level around batch $7$. PDE+RLOO
also improves early accuracy, achieving $42.4\%$ vs.\ $29.9\%$ at batch $1$,
$53.1\%$ vs.\ $36.0\%$ at batch $3$, and $54.6\%$ vs.\ $41.3\%$ at batch $4$.
The curves meet around batch $8$, after which RLOO finishes ahead. The advantage of PDE in this experiment is faster early learning.

\end{document}